%% file: 0_main.tex
\documentclass[11pt,a4paper]{article}
\usepackage[hyperref]{emnlp2020}
\usepackage{times}
\usepackage{latexsym}
\usepackage{graphicx}
\usepackage{amssymb}
\usepackage{amsmath}
\usepackage{enumitem}
\usepackage{pifont}

\usepackage{xcolor}

\usepackage{microtype}

\aclfinalcopy 
\def\aclpaperid{635} 

\title{Detecting Cross-Modal Inconsistency to Defend Against \\ Neural Fake News}

\author{Reuben Tan \\
  Boston University \\
  \texttt{rxtan@bu.edu} \\\And
  Bryan A. Plummer \\
  Boston University \\
  \texttt{bplum@bu.edu} \\ \And
  Kate Saenko \\
  Boston University \\
  \texttt{saenko@bu.edu} \\
  }

\date{}

\begin{document}
\maketitle
\begin{abstract}
Large-scale dissemination of disinformation online intended to mislead or deceive the general population is a major societal problem. Rapid progression in image, video, and natural language generative models has only exacerbated this situation and intensified our need for an effective defense mechanism. While existing approaches have been proposed to defend against \emph{neural fake news}, they are generally constrained to the very limited setting where articles only have text and metadata such as the title and authors. In this paper, we introduce the more realistic and challenging task of defending against machine-generated news that also includes images and captions. To identify the possible weaknesses that adversaries can exploit, we create a NeuralNews dataset composed of 4 different types of generated articles as well as conduct a series of human user study experiments based on this dataset. In addition to the valuable insights gleaned from our user study experiments, we provide a relatively effective approach based on detecting visual-semantic inconsistencies, which will serve as an effective first line of defense and a useful reference for future work in defending against machine-generated disinformation.
\end{abstract}

\input{1_introduction}

\input{2_related}
\input{7_data}
\input{8_userstudy}
\input{3_approach}
\input{4_experiments}
\input{5_conclusion}

\clearpage

\bibliographystyle{acl_natbib}
\bibliography{anthology,emnlp2020}

\clearpage
\onecolumn
\input{6_appendix}

\end{document}

%% file: 1_introduction.tex
\section{Introduction}
The rapid progression of generative models in both computer vision \cite{goodfellow2014generative,zhang2017stackgan,zhang2018self,choi2018stargan} and natural language processing \cite{jozefowicz2016exploring,radford2018improving, radford2019language} has led to the increasing likelihood of realistic-looking news articles generated by Artificial Intelligence (AI). The malicious use of such technology could present a major societal problem. \citet{zellers2019defending} report that humans are easily deceived by its AI-generated propaganda. By manipulating such technology, adversaries would be able to disseminate large amounts of online disinformation rapidly. While it is promising that the pretrained generative models themselves are our best defense \cite{zellers2019defending}, it is often challenging to be aware of the models utilized by adversaries beforehand. More importantly, it ignores the fact that news articles are often accompanied by images with captions \citep{lee2018stacked,ji2019saliency,huang2019few}. 


\begin{figure}
    \centering
    \includegraphics[width=\columnwidth]{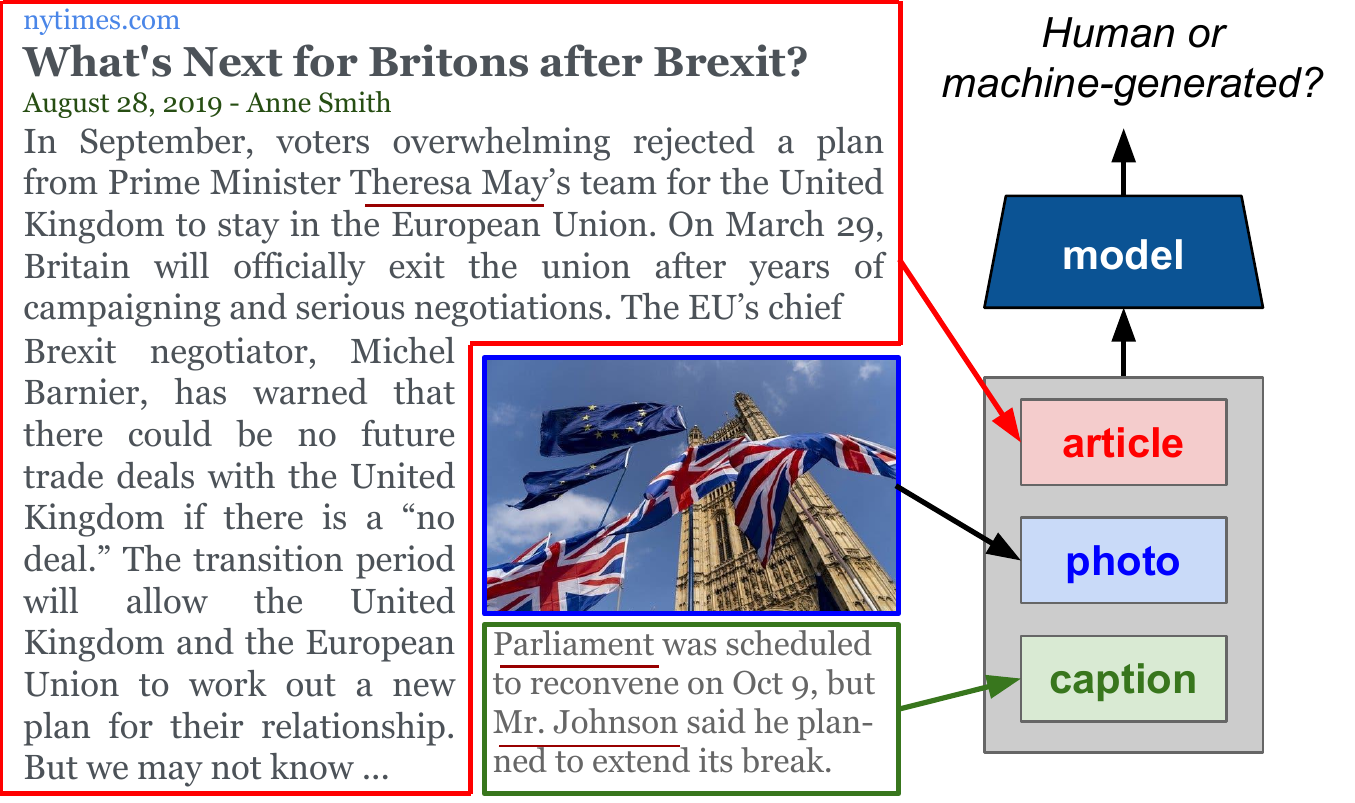}
    \caption{\small We propose an approach for detecting news articles generated by machines. Prior work uses only the article content and metadata including title, date, domain, and authors. However, news articles often contain photos and captions as well. We propose to leverage possible visual-semantic inconsistency between the article text, images, and captions, such as missing or inconsistent named entities (underlined in red). } \vspace{-5mm}
    \label{fig:motiv}
\end{figure}

In this paper, we present the first line of defence against \emph{neural fake news} with images and captions. To the best of our knowledge, we are the first to address this challenging and realistic problem. Premised on the assumption that the adversarial text generator is unknown beforehand, we propose to evaluate articles based on the semantic consistency between the linguistic and visual components. While state-of-the-art approaches in bidirectional image-sentence retrieval \citep{lee2018stacked,yang2019fast} have leveraged visual-semantic consistency to great success on standard datasets such as MSCOCO \cite{DBLP:journals/corr/LinMBHPRDZ14} and Flickr30K \cite{plummer2015flickr30k}, we show in Appendix~\ref{app:retrieval} they are not able to reason effectively about objects in an image and named entities present in the caption or article body. This is due to discrepancies in the distribution of these datasets, as captions in the standard datasets usually contain general terms including \emph{woman} or \emph{dog} as opposed to named entities such as \emph{Mrs Betram} and a \emph{Golden Retriever}, which are commonly contained in news article captions. Moreover, images are often not directly related to the articles they are associated with. For example, in Figure \ref{fig:motiv}, the article contains mentions of the British Prime Minister. Yet, it only contains an image of the United Kingdom flag. 

To circumvent this problem, we present DIDAN, a simple yet surprisingly effective approach which exploits possible semantic inconsistencies between the text and image/captions to detect machine-generated articles. For example, notice that the article and caption in Fig.~\ref{fig:motiv} actually mention different Prime Ministers. Besides evaluating the semantic relevance of images and captions to the article, DIDAN also exploits the co-occurrences of named entities in the article and captions to determine the \emph{authenticity score}. The \emph{authenticity score} can be thought of as the probability that an article is human-generated. We adopt a learning paradigm commonly used in image-sentence retrieval where models are trained to reason about dissimilarities between images and non-matching captions. In this instance, negative samples constitute articles and non-corresponding image-caption pairs. Not only is this a reasonable approach when the adversarial generative model is unknown, we show empirically that it is crucial to detecting machine-generated articles with high confidence even with access to machine-generated samples during training. More importantly, this means that DIDAN is easily trained on the abundance of online news articles without additional costly annotations.

To study this threat, we construct the NeuralNews dataset which contains both human and machine-generated articles. These articles contain a title, the main body as well as images and captions. The human-generated articles are sourced from the GoodNews \cite{biten2019good} dataset. Using the same titles and main article bodies as context, we use GROVER \cite{zellers2019defending} to generate articles. Instead of using GAN-generated images which are easy to detect even without exposure to them during training time \cite{wang2019detecting}, we consider the much harder setting where the articles are completed with the original images. We include both real and generated captions which are generated with the SOTA entity-aware image captioning model \cite{biten2019good}. We present results and findings from a series of empirical as well as user study experiments. In the user study experiments, we use 4 types of articles including real and generated news to determine what humans are most susceptible to. The insights derived from these findings help identify the possible weaknesses that adversaries can exploit to produce \emph{neural fake news} and serve as a valuable reference for defending against this threat. Last but not least, our experimental results provide a competitive baseline for future research in this area.

In summary,  our contributions are multi-fold:
\vspace{-2mm}
\begin{enumerate}
\itemsep0em
    \item We introduce the novel and challenging task of defending against full news article containing image-caption pairs. To the best of our knowledge, this is the first paper to address both the visual and linguistic aspects of defending against neural fake news.
    \item We introduce the NeuralNews dataset that contains both human and machine-generated articles with images and captions.
    \item We present valuable insights from our empirical and user study experiments that identify exploitable weaknesses.
    \item We propose DIDAN,  an effective named entity-based model that serves as a good baseline for defending against neural fake news. Most importantly, we empirically prove the importance of training with articles and non-matching images and captions even when the adversarial generative models are known.
\end{enumerate}

%% file: 2_related.tex
\section{Related Work}
\subsection{Neural News Generation and Defense}
GROVER \cite{zellers2019defending} draws on recent improvements in neural text generation \cite{jozefowicz2016exploring,radford2015unsupervised,radford2018improving,radford2019language} to generate realistic-looking articles complete with metadata such as title and publication date but without images. Interestingly, it also serves as the best form of defense against its own generated propaganda. \cite{adelani2020generating} show that the GPT-2 model can be manipulated to generate fake reviews to deceive online shoppers. Corroborating the findings by \cite{zellers2019defending}, they also report that pretrained language models such as GROVER and GPT-2 are unable to accurately detect fake reviews. To combat effectively against the dissemination of neural disinformation, \cite{tay2020reverse} propose a promising direction of reverse engineering the configurations of neural language models to identify detectable tokens. Last but not least, \cite{biten2019good} introduce an approach to generate image captions based on contextual information derived from news articles. Such progress points towards the inevitability of large-scale dissemination of generated propaganda and the significance of this task.

\subsection{Image and Video Generation and Defense} 
In recent years, the introduction of Generative Adversarial Networks \cite{zhang2017stackgan,zhang2018self,choi2018stargan} has led to unprecedented progress in image and video generation. While most of these have focused on generating images from text as well as video translation, such models can easily be exploited to generate disinformation which can be devastating to privacy and national security \cite{mirsky2020creation,chesney2019deep,chesney2019deepfakes}. In response to this growing threat, \cite{agarwal2019protecting} propose a forensic approach to identify fake videos by modeling people's facial expressions and speaking movements. In a similar vein to \cite{tay2020reverse}, \cite{matern2019exploiting, yang2019exposing,wang2019detecting,wang2019cnn} seek to exploit visual artifacts to detect face manipulations and deepfakes.  Encouragingly, \citet{wang2019detecting} show that neural networks can easily learn to detect generated images even without exposure to training samples from those generators.


%% file: 7_data.tex
\section{NeuralNews Dataset Collection} \label{data_collection}
To facilitate our endeavor of studying this threat, we introduce the NeuralNews dataset which consists of human and machine-generated articles with images and captions. It provides a valuable testbed for AI-enabled disinformation that adversaries can exploit presently and yet, is the hardest to detect.  The human-generated articles are sourced from the GoodNews \cite{biten2019good} dataset, which consists of New York Times news articles spanning from 2010 to 2018. Each news article contains a title, the main article body as well as image-caption pairs. Note that we source original images from real articles since machine-generated images are relatively easy to detect \cite{wang2019detecting}. In our dataset, we restrict the number of image-caption pairs to be at most 3 per article. The entire dataset used in the empirical and user study experiments contains the following 4 types of articles (see examples in Appendix~\ref{sec:examples}):
\begin{enumerate}[label=\Alph*]
\itemsep0em 
    \item) Real Articles and Real Captions
    \item) Real Articles and Generated Captions
    \item) Generated Articles and Real Captions
    \item) Generated Articles and Generated Captions
\end{enumerate}

In total, we collect about 32K samples of each article type (resulting in about 128K total). To collect machine-generated news articles, we use GROVER \cite{zellers2019defending} to generate fake articles using original titles and articles from the GoodNews dataset as context. Type C articles are completed by incorporating the original image-caption pairs. In Type B and D articles, we use the entity-aware image captioning model \cite{biten2019good} to generate fake image captions based on the articles. We believe that this dataset presents a realistic and challenging setting for defending against neural news. 

\smallskip

\noindent\textbf{Dataset Statistics.} Table \ref{tab:neural_news_stats} provides statistics on the length of articles and number of images in our Neural News dataset. Most articles contain at most 40 sentences in their main body. In addition, even though most articles contain a single image and caption, a sizeable 18.2\% have 3 images. We believe that this setting will provide a challenging testbed for future work to investigate methods using varying number of images and captions.

\begin{table}[t]
  \centering
  \small
  \setlength{\tabcolsep}{3.5pt}
  \begin{tabular}{|c|c|c|c|c|c|}
  \hline
    \# Sentences & \multicolumn{2}{|c|}{\% of Articles} & & \% of\\
    \cline{2-3}
    in Article & Real & Generated & \# Imgs &  Articles\\
    \hline
    $N \leq 10$ & 33.7 & 15.6 & 1 & 60.8\\
    $10 < N \leq 40$ & 54.4  &81.5 & 2 & 21.0\\
    $N > 40$ & 11.9 & 2.9 & 3 & 18.2\\
  \hline
  \end{tabular}
  \caption{\small NeuralNews dataset statistics across its 128K articles. Note that images are aggregated for both types of articles, since generated articles use the same images (but different articles and/or captions) as its corresponding real article.}
  \vspace{-3mm}
  \label{tab:neural_news_stats}
\end{table}



%% file: 8_userstudy.tex
\section{User Study Experiments} \label{sssec:human}
We endeavor to determine the susceptibility of humans to different types of neural fake news through several user studies. To this end, we conduct a series of user study experiments based on the NeuralNews dataset. The user study results provide vital information to help us identify salient points which adversaries can take advantage of. Qualified Amazon Mechanical Turk workers with required English proficiency are used in all of our experiments.  We briefly describe the experimental setups below. See Appendix~\ref{appendix: user} for a template of our prompts and response options for each setting.
\smallskip

\begin{figure}[t]
\begin{flushright}
\includegraphics[width=\columnwidth, height=6cm]{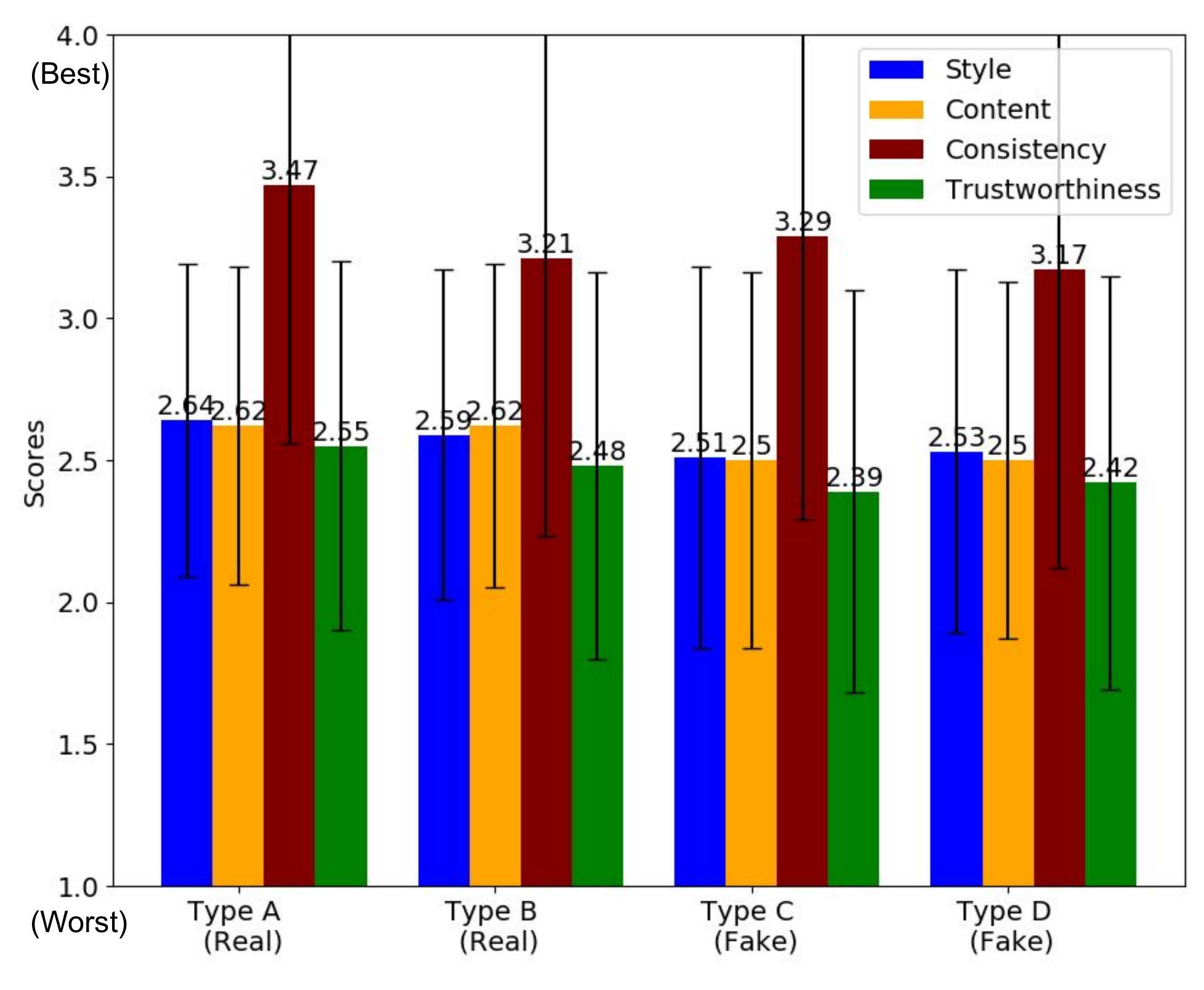}
\end{flushright}
\vspace{-3mm}
   \caption{\small \textbf{Trustworthiness results.} Human evaluation of the 4 article types in the trustworthiness experiment. Workers are asked to evaluate the article based on its style, content, consistency and overall trustworthiness. We observe that people generally have a hard time deciding on the overall trustworthiness on articles regardless of their types. The prompt and the response options can be found in Appendix \ref{appendix: user}.}
   \vspace{-5mm}
   
\label{fig:first_user_results}
\end{figure}

\begin{table*}[t]
  \centering
  \small
  \begin{tabular}{|c|c|c|c|c|c|}
  \hline
    Article &   &  & Article Only & Naive Participants & Trained Participants\\
   Type & Article & Caption & Accuracy & Accuracy & Accuracy\\
  \hline
    \hline
    no imgs & generated & -- & 68.8\% & -- & --\\
    no imgs & real & -- & 49.2\% & -- & --\\
    \hline 
    A & real & real & -- & 64.0\% & 70.7\%\\
    B & real & generated & -- & 34.0\% & 78.7\%\\
    C & generated & real & -- & 42.7\% & 56.7\%\\
    D & generated & generated & -- & 44.0\% & 55.3\%\\
    \hline
     Average & -- & - & 59.0\% & 46.2\% & 67.8\%\\
  \hline
  \end{tabular}
  \vspace{-2mm}
  \caption{\small \textbf{User prediction results.} We report the percentages of participants who are able to classify articles as human-generated or machine-generated accurately given different kinds of information and/or training (see Section~\ref{sssec:human} for additional details). A more in-depth breakdown of results can be found in Appendix~\ref{sec:user_study_breakdown}.} \vspace{-2mm}
  \label{tab:combined_user_results}
\end{table*}

\noindent\textbf{Trustworthiness: How well are humans able to rate the trustworthiness of news articles?} This experiment extends the study from~\citet{zellers2019defending} to also use images and captions. The goal is to understand the qualitative factors that humans consider to decide on the authenticity of articles by asking participants to evaluate articles based on style, content, consistency between text and images and overall trustworthiness using a four point scale where higher scores indicate more trust. 
\smallskip
  
\noindent\textbf{Article Only User Predictions: Given articles with titles which do not contain images and captions, can humans detect if they are generated or not?}We ask participants to predict if an article is fake when they only contain their titles and main article bodies. In this variant, participants are provided with hints to pay attention to possible inconsistencies between the text and title. This is done with the purpose of understanding the importance of visual-semantic cues provided by image-caption pairs in this task in the following experiments.
\smallskip
  
\noindent\textbf{Naive User Predictions: Can humans discern if an article is real or generated without prior exposure to generated articles?} In this experiment, participants are tasked to decide based on their own judgements if the articles are human or machine-generated after reading them. The intuition behind this experiment is to determine humans' capability to identify fake news without prior exposure.
\smallskip
    
\noindent\textbf{Trained User Predictions: Are humans able to detect generated articles if they are told what aspects to pay attention to beforehand?} We provide limited training to participants by showing them examples of human and machine-generated articles that specifically highlighted semantic inconsistencies between articles and image-caption pairs.  Afterwards, we ask our trained participants to determine if a article is human or machine generated as done for the naive user predictions.

\subsection{User Study Results}
Figure \ref{fig:first_user_results} reports the results from our trustworthiness experiment where participants evaluated the overall trustworthiness of the article, but were not asked to determine if it was real or machine generated. These results show that humans generally have trouble with agreeing on the semantic relevance between images and the text (article body and captions), as evident from the large variance in their responses. We hypothesize the \emph{loose} connection between an article and an image (\ref{fig:motiv} to be a possible factor. Adversaries could easily exploit this to disseminate realistic-looking neural fake news. Consequently, exploring the visual-semantic consistency between the article text and images could prove to be an important area for research in defending against generated disinformation. While it is reassuring that the overall trustworthiness for the human-generated articles is the highest among the different article classes, these results also highlight the susceptibility of humans to being deceived by generated neural disinformation. The difference between the overall trustworthiness ratings across the different article classes is marginal.

Table \ref{tab:combined_user_results} reports the aggregated percentages of participants who are able to detect human and machine-generated articles accurately from the rest of the user study experiments. See Appendix \ref{sec:user_study_breakdown} for a complete breakdown of results. Trained participants are deemed to have classified a Type B article correctly if they select any of 4 responses that indicate visual-semantic inconsistencies between images and the article or captions. The significant difference in the detection accuracy of Type B articles between the naive and trained users suggest that humans do not typically pay much attention to image captions in online news. However, it is also reassuring that 14\% more participants are able to detect them after prior exposure.

We predict that Type C articles will be the most likely type of neural disinformation that adversaries would exploit for their purposes, given the current state of SOTA neural language models and image-captioning models. While recent neural language models are able to produce realistic-looking text, SOTA image captioning models are generally not able to generate captions of comparable quality. Oftentimes, the generated captions contain repeated instances of named entities without any stop words.

In summary, it is worrying that humans are particularly susceptible to being deceived by Type C and D articles in Table \ref{tab:combined_user_results}. However, we believe that there are fewer repercussions from the spread of Type B articles with real article content and generated captions. Since the generated captions only makes up a very small component of the entire article, the information conveyed is less likely to mislead people. In contrast, Type C articles have the potential to be exploited by adversaries to disseminate large amount of misleading disinformation due to its generated article contents. Consequently, our proposed approach is geared towards addressing this particular type of generated articles.

%% file: 3_approach.tex
\section{\normalfont \textbf{DIDAN:} \textbf{D}etecting Cross-Modal \textbf{I}nconsistency to \textbf{D}efend \textbf{A}gainst \textbf{N}eural Fake News}
\begin{figure*}[t]
\begin{center}
\includegraphics[width=\linewidth]{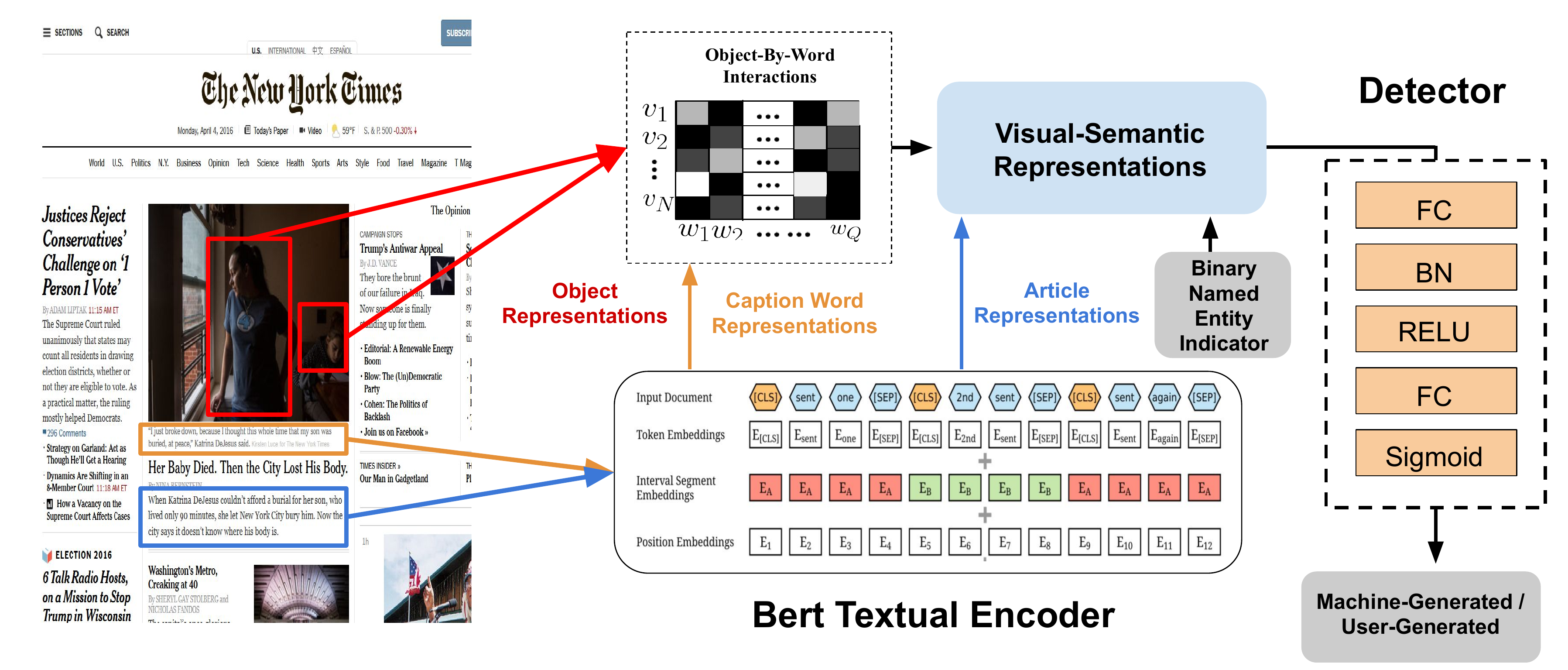}
\end{center}
   \caption{\small An overview of our proposed DIDAN model. To reason about relationships between named entities present in the article and entities in an image, DIDAN integrates article context into the visual-semantic representation learned from fine-grained object-by-word interactions. The aforementioned visual-semantic representation is used to infer an authenticity score for the entire news article.}
\label{fig:model}
\end{figure*}
In our task, the goal is to detect machine-generated articles that also include images and captions. The example in Figure \ref{fig:motiv} points towards an inherent challenge: identifying indirect relationships between the image and the text. Due to the common need to measure visual-semantic similarity, an intuitive first step would be to base one's approach on image-sentence similarity reasoning models which are commonly used in SOTA bidirectional image-sentence retrieval. We hypothesize, from their dismal performance (Table \ref{tab:retrieval}), that the image-sentence retrieval models are not adept at relating named entities in the articles to objects in the images. This suggests  that contextual information about named entities from the article body is essential.

As a first line of defense, we present our named entity-based approach DIDAN. Besides integrating contextual information from the text, DIDAN factors in the co-occurrence of named entities in the article body and caption to detect possible visual-semantic inconsistencies. This is based on a simple observation that captions, more often than not, contain mentions of named entities that are present in the main body too. DIDAN is trained on real and generated articles. To train our model to detect visual-semantic inconsistencies between images and text, we also adopt the learning paradigm from image-sentence similarity models. In this case, the negative samples are real but the article and its image-caption pairs are mismatched.


An illustrative overview of DIDAN is shown in Figure \ref{fig:model}. An article $A$ consists of a set of sentences $S$ where $S = \{S_1, \cdot \cdot \cdot, S_A\}$. Each sentence $S_i$ contains a sequence of words $\{w_1, \cdot \cdot \cdot, w_i\}$. The article is also comprised of a set of image-caption pairs where each image $I$ is represented by a set of regional object features $\{o_1, \cdot\cdot\cdot, o_I\}$ and each caption $C$ contains a sequence of words $C = \{w_1, \cdot \cdot \cdot, w_I\}$. Spacy's named entity recognition model \cite{honnibal2017spacy} is used to detect named entities in both articles and image captions. $d^T$, $d^I$ and $d^{vse}$ are used to denote the initial dimensions of the text and image representations as well as the hidden dimension respectively. Each sentence is tokenized and encoded with a BERT model \cite{devlin2018bert} that is pretrained on BooksCorpus \cite{zhu2015aligning} and English Wikipedia.

\subsection{Article Representations} 
\label{art_rep}
To extract relevant semantic context from the article, we begin by computing sentence representations. For each sentence $S^i$ in article $A$, the word representations are first projected into the article subspace as follows:
\begin{equation}
    S^i = W^{art}V^i
\end{equation}
where $V^i$ represent all word embeddings in $S^i$. For a given sentence $S^i$, its representation $S^i_f$ is computed as the average of all its word representations where the subscript $f$ denotes the corresponding representation. In turn, the article representation $A_f$ for an article A is computed as the average of all its sentence representations.

\subsection{Visual-Semantic Representations}
Our approach leverages word-specific image representations learned from images and captions to determine their relevance to an article. A caption is represented by a feature matrix $V^{cap}_f \in \mathbb{R}^{n_c \times D^T}$ and an image is represented by a matrix of object features $V^{vis}_f \in \mathbb{R}^{n_o \times D^{I}}$. 
As in the previous section, the word embeddings of a caption and image object features are projected into a common visual-semantic subspace using:
\begin{equation}
    C_f^{cap} = W^{cap}V^{cap}_f
\end{equation}
\begin{equation}
    I_f^{vis} = W^{vis}V^{vis}_f
\end{equation}
A key property of these visual-semantic representations is that they are built on fine-grained interactions between words in the caption and objects in the image. To begin, a semantic similarity score is computed for every possible pair of projected word and object features $w_l$, $v_k$, respectively.
\begin{equation}
    \centering
    s_{kl} = \frac{v_k^{T} w_l}{\left \| v_k \right \| \left \| w_l \right \|} \text{ where }  k \in [1, n_o] \text{ and } l \in [1, n_c].
    \label{eq:obw_matrix}
\end{equation}
where $n_c$ and $n_o$ indicate the number of words and objects in a caption and image, respectively.
These similarity scores are normalized over the objects to determine the salience of each object with respect to a word in the caption.
\begin{equation}
    \centering
    a_{kl} = \frac{\exp(s_{kl})}{\sum_{i=1}^{n_o} \exp(s_{il})}.
    \label{eq:word_attend}
\end{equation}
The word-specific image representations are computed as a weighted sum of the object features based on the normalized attention weights:
\begin{equation}
    \centering
   w_l^I = a_l^T I_f^{vis}
    \label{eq:frame_attend}
\end{equation}

\subsection{Detector}
A key contribution of our approach is the utilization of a binary indicator feature, which indicates if the caption contains a reference to a named entity present in the main article body. The article representation and the average of the word-specific image representations are concatenated to create caption-specific article representations which are passed into the discriminator:
\begin{equation}
    \centering
    A^c_f = concat\{A_f, \frac{1}{n_c}\sum_{l=1}^{n_c}w_{l}^I, b_c\}
\end{equation}
where $concat\{\cdot\cdot\cdot\}$ denotes the concatenation operation and $b_c$ is the binary indicator feature. The key insight is that article context is integrated into the caption-specific article representations. Our discriminator (Figure \ref{fig:model}) is a simple neural network that is comprised of a series of Fully-Connected (FC),  Rectified Linear Unit (ReLU), Batch Normalization (BN) and sigmoid layers. It outputs an \emph{authenticity} score for every image-caption pair. 

Recall that in our problem formulation news articles can contain varying numbers of images and captions. The final \emph{authenticity score} of an article is determined across those of its images and captions. It can be thought of as the probability that an article is human-generated. The \emph{authenticity score} is computed across the set of images and captions in an article as follows:
\begin{equation}
    \centering
    p_A = 1 - \prod_{images}(1 - p_A^I) 
\end{equation}
where $p_A^I$ is the \emph{authenticity score} of image-caption pair I with respect to article A. Intuitively, if an image-caption pair is deemed to be relevant to the article body (scores close to 1), then the final \emph{authenticity score} will be close to 1 as well.

The entire model is optimized end-to-end with a binary cross-entropy loss. 
\begin{equation}
\begin{split}
L =  - \sum_{(A^+, I^+)} \sum_{I^-} y\log(p_A) + (1-y)log(1-p_A).
\end{split}
\label{eq:total_loss}
\end{equation}
where $I^-$ denotes non-matching sets of images and captions with respect to an article and y is the ground-truth label of an article. Negative images and their captions are sampled from other articles within the same minibatch. 

%% file: 4_experiments.tex
Given a news article from our NeuralNews dataset, our goal is to automatically predict whether it is human or machine-generated. We compare DIDAN to several baselines, evaluating performance based on how often an article was correctly labeled. Note that in our experiments, only Type A and C articles are used. This is due to the fact that generated captions often contain repeated instances of named entities without any stop words, which is not challenging for humans to detect (see Table~\ref{tab:combined_user_results}). To comprehend the importance of each component of DIDAN and each part of the news article, we supplement our analysis with ablation experiments.

\begin{table*}[t]
  \small
  \centering
  \begin{tabular}{|c|c|c|c|c|c|}
  \hline
    Approach & Trained With &  Named Entity & Generated Articles & GROVER-Mega & GROVER-Large\\
    & Mismatch & Indicator & in Training (\%) & Accuracy (\%) & Accuracy (\%)\\
    \hline \hline
     CCA &  - & - & None & 52.1 & -\\
    \hline
    DIDAN &\checkmark  & - & None & 54.5 & -\\
      &\checkmark  &  \checkmark & None & {\bf 64.1} & -\\
    \hline \hline
     Grover Discriminator &  - & - & 50 & 56.0 & -\\
    \hline
    & - & - & 25 & 51.2 & 49.9\\
    & - & - & 50 & 56.4 & 53.7\\
    \cline{2-6}
    & - &  \checkmark & 25 & 64.9 & 64.6 \\
    DIDAN & - &  \checkmark & 50 & 68.8 & 66.3\\
    \cline{2-6}
    &\checkmark & - & 25 & 61.0 & 65.0\\
    &\checkmark & - & 50 & 70.3 & 57.4\\
    \cline{2-6}
    &\checkmark  &  \checkmark & 25 & 80.9 & 69.8\\
    &\checkmark  &  \checkmark & 50 & {\bf 85.6} & {\bf 77.6}\\
  \hline
  \end{tabular}
  \caption{\small Results of machine-generated (with GROVER-Mega) vs real news detection on our NeuralNews dataset. We show performance of DIDAN variants trained on generated (with GROVER-Large or GROVER-Mega) articles and image-captions pairs when the number of generated articles is limited during training time. \emph{Mismatch} indicates real data but with images and captions that do not correspond to the article body. The percentages of real and machine-generated articles do not change across variants that are trained with or without \emph{mismatch} data.}
  \label{tab:combined_results}
\end{table*}

\begin{table*}[ht]
  \centering
  \small
  \begin{tabular}{|c|c|c|c|c|c|}
  \hline
    Articles & Images & Captions & DIDAN Accuracy (\%) & CCA Accuracy (\%)\\
   \hline
   \checkmark & \checkmark & \checkmark & 85.6 & 51.4 \\
   \checkmark & - & \checkmark & 81.9 & 50.1 \\
   \checkmark & \checkmark & - & 56.9 & 52.1 \\
  \hline
  \end{tabular}
  \caption{\small Ablation results of CCA and DIDAN's detection accuracy w.r.t. the contribution of each component in the news article. Experiments are performed on NeuralNews and the training as well as testing articles are generated by GROVER-Mega.}
  \label{tab:cca_results}
  \vspace{-5pt}
\end{table*}

\subsection{Implementation Details and Baselines}
Our model is implemented using Pytorch. In our implementation, the dimensions for the Bert-Base and object region features $d^T$ and $d^I$ are set to 768 and 2048 respectively. We also set the dimension of the visual-semantic embedding space $d^{vse}$ to be 512. The image region representations are extracted with the bottom-up attention \cite{Anderson2017up-down} model that is pretrained on Visual Genome \cite{krishna-etal-2017-dataset}. The language representations are extracted from a pretrained BERT-Base model \cite{devlin2018bert}. We adopt an initial learning rate of $1e^{-3}$ and train our models end-to-end using the ADAM optimizer. 

In addition to ablations of our model, we also compare to a baseline using Canonical Correlation Analysis (CCA), which learns a shared semantic space between two sets of paired features, as well as the GROVER Discriminator.  In our CCA implementation, images are represented as the average of its object region features and the caption is represented by the average of its word features. We apply CCA between the article features (Section \ref{art_rep}), and the concatenation of the image and caption features. The projection matrices in CCA are learned from positive samples constituting articles and their corresponding images and captions. The GROVER Discriminator is a simple neural network used in \cite{zellers2019defending} to detect its own generated articles based on the article text and metadata. We train the GROVER Discriminator without mismatched data and without images or captions. 
\noindent\textbf{Training on Real News Only.} The top of Table~\ref{tab:combined_results} shows that our approach significantly improves  detection accuracy when trained without any generated examples (i.e. with mismatched real news as negatives) compared to CCA. Our named entity indicator (NEI) features provide a large improvement in this most difficult setting.
\smallskip

\noindent\textbf{Training with Generated Samples.} 
We consider the realistic setting where generated articles may be available but the generator is not. We report the performance achieved by variants of DIDAN when trained on Grover-Mega generated articles in the second-to-last column of Table~\ref{tab:combined_results}. Note that the result achieved by GROVER Discriminator, akin to our text-only variant, is substantially worse than the result reported in \cite{zellers2019defending}. This is because we train it with BERT representations as opposed to leveraging GROVER learned representations to detect its own generated articles.  Based on the consistent trend of the results, training on generated articles from the same generator as appears in test data improves the capability of a neural network to detect them. The binary NEI features also prove to be very beneficial to increasing the detection accuracy of DIDAN. Interestingly, even when we have access to generated articles during training, the large improvement in detection accuracy going from 68.8\% 
to 85.6\% when also training on mismatched \textit{real} images and captions suggests that visual-semantic consistency has an important role to play in defending against neural fake news.
\smallskip

\noindent\textbf{Unseen Generator.} To evaluate DIDAN's capability to generalize to articles created by generators unseen during training, we train on GROVER-Large generated articles and evaluate on GROVER-Mega articles (last column of Table~\ref{tab:combined_results}). While overall accuracy drops, we observe the same trend where our proposed training with mismatched real data helps increase the detection accuracy from 66.3\% to 77.6\%, and removing NEI lowers accuracy.
\smallskip

\noindent\textbf{Images vs Captions.} Table \ref{tab:cca_results} show more ablation results of our model and CCA on NeuralNews. We observe an improvement of 2\% in accuracy achieved by CCA variants with images. This suggests that visual cues from images can provide contextual information vital to detecting generated articles. This is also corroborated by the ablation results obtained by DIDAN, where we observe that both images and captions are integral to detecting machine-generated articles. While the contribution of the captions is the most significant, we note that the visual cues provided by images are integral to achieving the best detection accuracy.

\subsection{Visualizations}
\begin{figure*}[t]
    \centering
    \includegraphics[width=\linewidth, height=8cm]{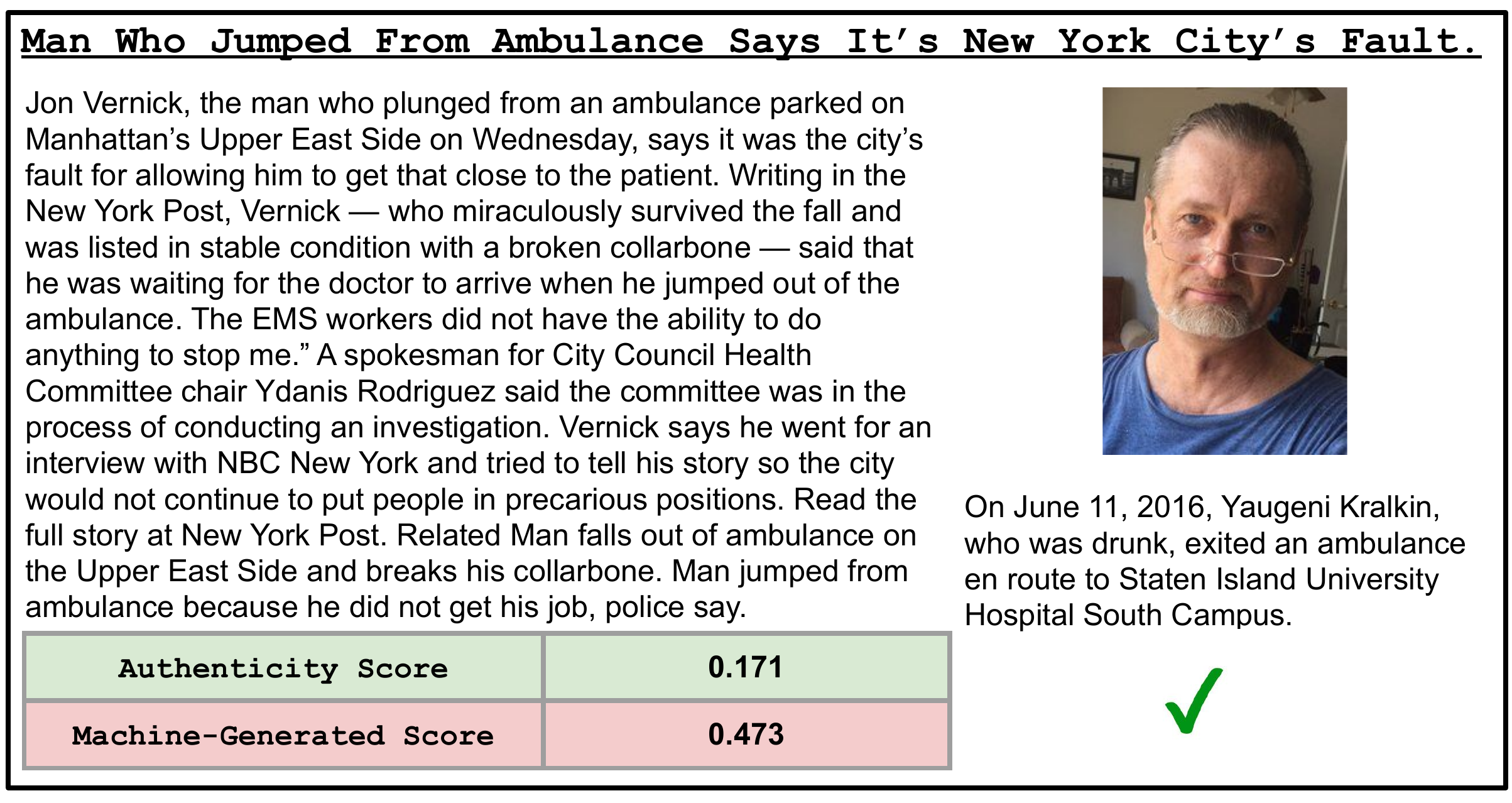}
    \caption{A machine-generated article that was classified correctly as such by DIDAN.}
    \label{fig:visualization_1}
    \vspace{-10pt}
\end{figure*}

\begin{figure*}[ht!]
    \centering
    \includegraphics[width=\linewidth, height=8cm]{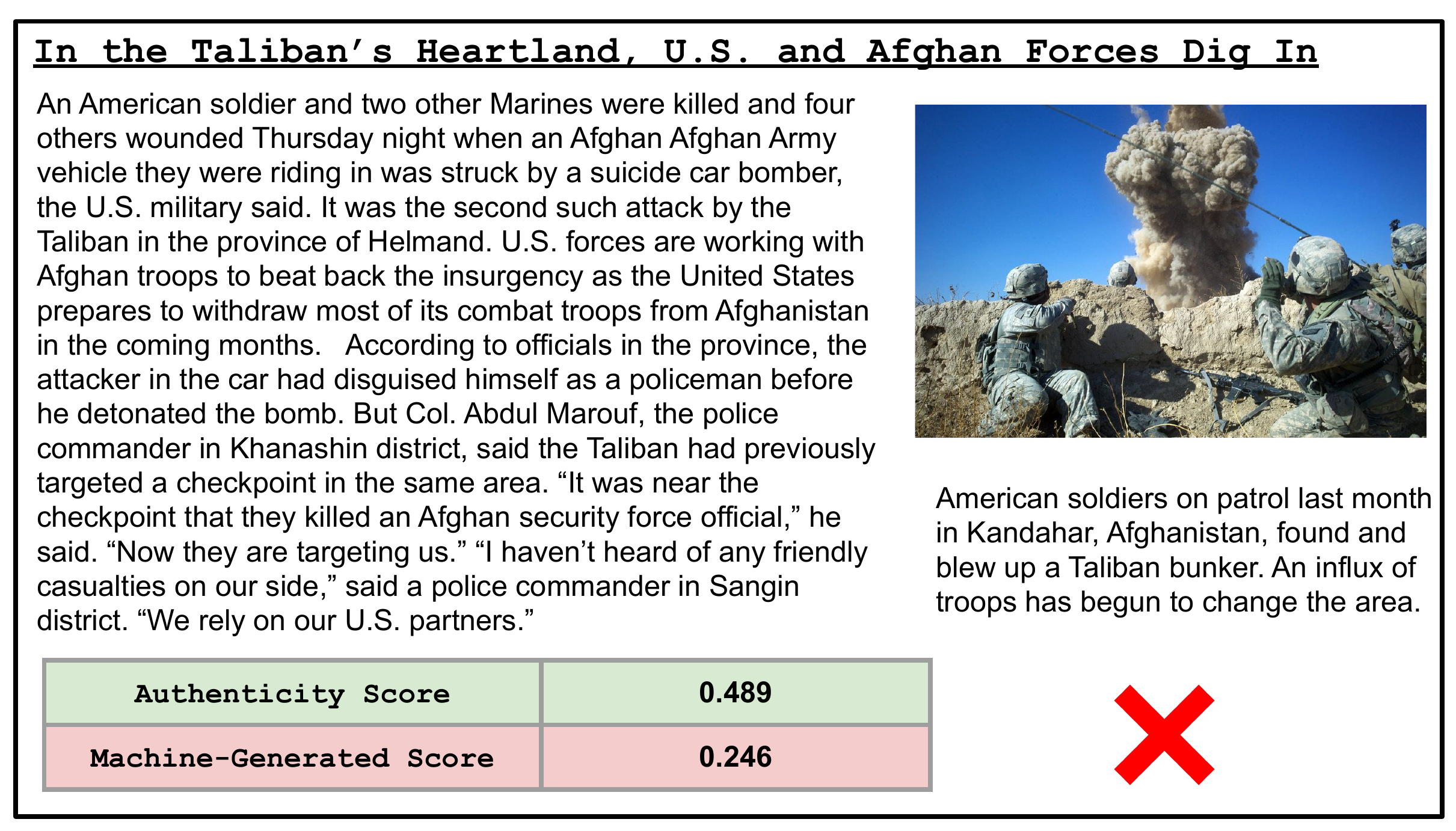}
    \caption{A machine-generated article that was classified incorrectly by DIDAN.}
    \label{fig:visualization_2}
    \vspace{-7pt}
\end{figure*}

In Figure \ref{fig:visualization_1} and \ref{fig:visualization_2} we present examples of our model's prediction of sample articles (additional examples can be found in appendix \ref{appendix_visualizations}). In Figure \ref{fig:visualization_1} we observe that DIDAN is able to classify a machine-generated article correctly. One plausible reason for this is that the main subject in the caption does not match the person who was mentioned in the article body and DIDAN is able to pick up on this relatively easily. However, the example in Figure \ref{fig:visualization_2} presents an especially challenging setting for DIDAN. In this case, the caption is only loosely related to the article and the image may or may not portray the situation described in the article.  Successfully determining the relevance of such relationships requires more abstract reasoning, which may be a good direction for future work.

%% file: 5_conclusion.tex
\section{Summary Of Exploitable Weaknesses and Defense Directions}
While this is not entirely representative of all the future challenges presented by \emph{neural fake news}, we believe that this comprehensive study will provide an effective initial defense mechanism against articles with images and captions. Based on the findings from the user evaluation, humans may be easily deceived by articles generated by SOTA models if they are not attuned to noticing possible visual-semantic inconsistencies between the article text and images. Adversaries can easily exploit this fact to create misleading disinformation by generating fake articles and combining them with manually sourced images and captions.

Encouragingly, our experimental results suggest that visual-semantic consistency is an important and promising research area in our defense against neural news.

We hope future work will address any potential limitations of this work, such as expanding the dataset to evaluate generalization across different news sources, and a larger variety of neural generators. Other interesting avenues for future research is to understand the importance of metadata in this multimodal setting and investigating counter-attacks to improved generators that incorporate image-text consistency. Last but not least, DIDAN and NeuralNews may be leveraged to supplement fact verification in detecting human-written misinformation in general by evaluating visual-semantic consistency. 

\noindent\textbf{Acknowledgements:} This work is supported in part by NSF awards IIS-1724237, CNS-1629700, CCF-1723379. We would also like to thank Professor Derry Wijaya for the valuable discussions and feedback.

%% file: 6_appendix.tex
\appendix
\maketitle
\thispagestyle{empty}

\section{User Study Templates} \label{appendix: user}

\subsection{Trustworthiness Study Template} 
This experiment requires users to evaluate the quality of news articles based on the following 4 criteria. The response options are displayed next to their corresponding score ratings. For ease of comparisons, we adopt the same metrics and scoring system in \cite{zellers2019defending}.
\begin{enumerate}[label=(\alph*)]
    \item (Style) Is the style of the article consistent?
    \begin{enumerate}[label=\arabic*)]
        \item No, it reads like it’s written by a madman.
        \item Sort of, but there are certain sentences that are awkward or strange.
        \item Yes, this sounds like an article I would find at an online news source.
    \end{enumerate}
    \item (Content) Does the content of this article make sense?
    \begin{enumerate}[label=\arabic*)]
        \item No, I have no (or almost no) idea what the author is trying to say.
        \item Sort of, but I don’t understand what the author means in certain places.
        \item Yes, this article reads coherently.
    \end{enumerate}
    \item (Consistency) Does the text match the images?
    \begin{enumerate}[label=\arabic*)]
        \item No, the images do not match the text and captions.
        \item Sort of, the images match the captions but do not match the text.
        \item Sort of, the images match the text but do not match their captions.
        \item Yes, the images match the text and captions.
    \end{enumerate}
    \item (Trustworthiness) Does the article read like it comes from a trustworthy source?
    \begin{enumerate}[label=\arabic*)]
        \item No, this seems like it comes from an unreliable source.
        \item Sort of, but something seems a bit fishy.
        \item Yes, I feel that this article could come from a news source I would trust.
    \end{enumerate}
\end{enumerate}

\subsection{Naive User Predictions Study Template} \label{appendix: second_user}
In the second user study experiment, users are asked to decide based on their own judgements if the articles are human or machine-generated after reading them. These articles contain images and captions.
\begin{enumerate}[label=(\alph*)]
    \item Do you think this article is human or machine-generated?
     \begin{enumerate}[label=\arabic*)]
        \item Human-generated.
        \item Machine-Generated.
    \end{enumerate}
\end{enumerate}

\subsection{Trained User Predictions Study Template} \label{appendix: third_user}
In this variant, users are given hints to pay more attention to specific components of news articles through the provided response options. The response options provide users with cues to look at possible semantic inconsistencies between the articles and image-caption pairs.
\begin{enumerate}[label=(\alph*)]
    \item Choose the rating that you think is the most suitable for the given news article.
     \begin{enumerate}[label=\arabic*)]
        \item Human-Generated.
        \item Machine-generated because 1 or more images is not very relevant to the article body.
        \item Machine-generated because 1 or more captions is not very relevant to the article body.
        \item Machine-generated because 1 or more images are not very relevant to the caption.
        \item Machine-generated because the image/caption pairs are not relevant to the article body.
        \item Machine-generated because something about the article appears off.
        \item Machine-generated because the article title is not really relevant to the article.
    \end{enumerate}
\end{enumerate}

\subsection{Article-only User Predictions Study Template}
The fourth user study experiment is similar to that of the third experiment except that it does not contain images or captions. Instead, users are told to focus on possible mismatches between the title and article body.

\begin{enumerate}[label=(\alph*)] \label{appendix: fourth_user}
    \item Choose the rating that you think is the most suitable for the given news article.
     \begin{enumerate}[label=\arabic*)]
        \item Human-generated.
        \item Machine-generated because something about the article appears off.
        \item Machine-generated because the article title is not really relevant to the article.
    \end{enumerate}
\end{enumerate}

\section{User Study Results} \label{sec:user_study_breakdown}

\subsection{Naive User Study Results}
\begin{table}[h]
  \centering
  \begin{tabular}{|c|c|c|}
  \hline
    Article Type & Accuracy \\
    \hline
    A & 64.0\%\\
    B & 34.0\% \\
    C & 42.7\%\\
    D & 44.0\% \\
    \hline
    Total & 46.2 \%\\
  \hline
  \end{tabular}
  \caption{Results of the naive user predictions experiment. In this study, workers rely on their own judgement to decide if articles are human or machine-generated after reading them. The results present a worrying trend where a majority of the workers are misled by generated neural disinformation. The prompt and the response options can be found at \ref{appendix: second_user}.}
  \label{tab:second_user_results}
\end{table}

\noindent The findings in Figure \ref{fig:first_user_results} are corroborated by the results from the naive user prediction study in Table \ref{tab:second_user_results}. The lower than random classification accuracy of 46.2\% suggests that discriminating between human and machine-generated articles is a very challenging task in general. In particular, it is worrying that only 42.7\% of users are able to accurately identify the Type C articles comprised of generated article bodies and real image-caption pairs.

\subsection{Trained User Study Results}

\begin{table*}[h]
  \centering
  \begin{tabular}{|c|c|c|c|c|c|c|c|c|}
  \hline
  Article Type & 1 & 2 & 3 & 4 & 5 & 6 & 7 & Accuracy\\
  \cline{3-9}
    \hline
    A & 70.7 & 7.3 & 6.0 & 4.0 & 4.7 & 2.7 & 2.7 & 70.7\\
    B & 11.3 & 52.0 & 8.0 & 14.0 & 4.7 & 8.7 & 1.3 & 78.7\\
    C & 43.3 & 13.3 & 11.3 & 10.7 & 8.0 & 1.3 & 12.0 & 56.7\\
    D & 44.7 & 12.7 & 14.0 & 13.3 & 8.7 & 5.3 & 1.3 & 55.3\\
    \hline
    Average & - & - & - & - & - & - & - & 67.8\\
  \hline
  \end{tabular}
  \newline
  \caption{Results of the trained user predictions experiment. In this study, workers are prompted to pay more attention to specific aspects of the articles by the response options before selecting the most appropriate response. The values in the columns with numerical headings indicate the percentage of users who select the corresponding response for each class of article. Generally, rating 1 indicates that the article is human-generated and the rest indicate otherwise due to possible semantic irrelevance between the articles, images and captions. The prompt and the exact rating descriptions can be found at Appendix \ref{appendix: third_user}.}
  \label{tab:third_user_results}
\end{table*}

\noindent In the trained user prediction study, users are provided with hints to focus on possible visual-semantic inconsistencies between the article text (main body and image captions) and images via the provided response options. Table \ref{tab:third_user_results} reports the percentage of users who selected each response for the different classes of articles. The numerical headings in Table \ref{tab:third_user_results} indicate their corresponding responses as shown in appendix \ref{appendix: third_user}. We observe a recurring theme where a large percentage of users are deceived by Type D articles. Only 55.3\% of users identified the aforementioned article class as generated. It is also notable that workers who are told to focus on possible visual-semantic inconsistencies are significantly more accurate in detecting generated articles. 

\subsection{Article-Only User Study Results}
\begin{table*}[h]
  \centering
  \begin{tabular}{|c|c|c|c|c|}
  \hline
    Article Type & 1 & 2 & 3 & Accuracy\\
    \hline
    Human-Generated & 49.2 & 36.4 & 14.4 & 49.2\\
    Machine-Generated & 31.2 & 61.6 & 7.2 & 68.8\\
    \hline
    Average & - & - & - & 59.0\\
  \hline
  \end{tabular}
  \caption{Results of the article-only user predictions experiment. This study is similar to the trained user prediction study. However, in this experiment, the sample articles do not contain any image-caption pairs. Instead, each article sample only contains a title and the main body. The values in the columns with numerical headings indicate the percentage of users who select the corresponding response for each class of article. The prompt and the response options can be found at \ref{appendix: fourth_user}.}
  \label{tab:fourth_user_results}
\end{table*}

\noindent The results from the article-only user study are reported in Table \ref{tab:fourth_user_results}. In this experiment, workers are provided with hints to focus on possible semantic inconsistencies between the title and main body. The articles do not contain image-caption pairs. It is observed that by focusing on possible semantic inconsistency between the title and article body, a large majority of workers are able to identify generated articles.


\section{Importance of metadata in GROVER}
\begin{table*}[h]
  \centering
  \begin{tabular}{|c|c|c|c|c|c|c|c|c|}
  \hline
  Article & Authors & Date & Domain & Title & Bert-Large Accuracy & Pretrained Grover Accuracy\\
  \cline{3-7}
    \hline
    \checkmark & - & - & - & - & 73.0 & 81.6\\
    \checkmark & \checkmark & \checkmark & \checkmark & - & 76.8 & 90.0\\
    \checkmark & \checkmark & \checkmark & - & \checkmark & 75.2 & 90.3\\
    \checkmark & \checkmark & - & \checkmark & \checkmark & 71.7 & 90.0\\
    \checkmark & - & \checkmark & \checkmark & \checkmark & 69.6 & 90.6\\
    \checkmark & \checkmark & \checkmark & \checkmark & \checkmark & 70.8 & 90.1\\
  \hline
  \end{tabular}
  \newline
  \caption{Ablation results of our model and the pretrained Grover model on the Grover \cite{zellers2019defending} discrimination dataset.}
  \label{tab:our_grover_results}
\end{table*}

\noindent We present results from a series of ablation experiments on the metadata which include the authors, date, domain and title. The ablation experiments are performed on the Grover discrimination dataset. Table \ref{tab:our_grover_results} report results from ablation experiments achieved by a BERT-Large model and Grover on its discriminated dataset. While using metadata generally leads to increased accuracy in detecting generated articles across both models, the resulting improvement is more pronounced on the Grover model. Despite the fact that leveraging metadata significantly improves the performance of Grover, it also appears that the accuracy does not vary much with the exclusion of different types of metadata. In contrast, we observe a surprising observation that leveraging all metadata causes the detection accuracy to decrease. In addition, the inclusion of title results in a 6\% decrease in detection accuracy. Without knowledge of the adversary's generative language model, it is essential to understand the contribution of such metadata in defending against general neural disinformation.

\clearpage

\section{Bidirectional Image-Sentence Retrieval Results}
\label{app:retrieval}
\begin{table*}[h]
  \centering
  \begin{tabular}{|c|c|c|c|c|c|}
  \hline
    Variants & Directions & R@1 & R@5 & R@10 & Average\\
    \hline
    SCAN & Image to Text & 0.1 & 0.6 & 1.1 & 0.6\\
    \cline{2-6}
     & Text to Image & 0.1 & 0.5 & 1.0 & 0.5\\
    \hline
    SCAN + & Image to Text & 1.0 & 16.5 & 23.0 & 13.5\\
    \cline{2-6}
    NER + Face Recognition & Text to Image & 2.2 & 6.5 & 9.2 & 6.0\\
  \hline
  \end{tabular}
  \caption{Bidirectional Image-Sentence Retrieval Results obtained on images and captions from the GoodNews dataset.}
  \label{tab:retrieval}
\end{table*}

We observe that standard image-sentence retrieval models perform really badly on images and captions extracted from the GoodNews \cite{biten2019good} dataset. We hypothesize that image-sentence retrieval models are designed to measure visual-semantic similarity between images and phrases that contain general terms such as \emph{man} or \emph{dog}. In contrast, they are less capable of reasoning about relationships between images and named entities often found in news captions.

\section{Examples of Article Types}\label{sec:examples}
We provide samples of the different types of articles below. Each article sample contains a title, text, an image and a caption. The image and caption are located below the article text.

\subsection{Type A Article}
\bigskip
\centerline{\textbf{\Large Playing Composer, of Course, to Impress}}
\bigskip

At a time when opportunities for gifted emerging opera composers blazing all manner of new stylistic
trails appear to be proliferating, there's something to be said for a company willing to go to bat for
fresh pieces by veteran creators working in conventional modes. Not long ago, that company likely
would have been the Dicapo Opera, which performed an estimable service in championing
composers like Thomas Pasatieri, Tobias Picker and Conrad Susa. But with Dicapo in a state of
limbo, it falls to other institutions to fill the void. Kudos, then, to the Bronx Opera Company, which
opened its 47th season on Saturday night with "The Rivals," a 2002 comic opera by Kirke Mechem,
in the Lovinger Theater at Lehman College. Mr. Mechem, born in Wichita, Kan., and based in San
Francisco at 88, is a skillful composer especially admired for his vocal music. "Tartuffe," his first
opera, has played more than 350 times since its 1980 San Francisco Opera premiere. Mr. Mechem
fashioned his own libretto for "The Rivals," his third opera, relocating an 18th-century Sheridan
comedy from Bath, England, to Newport, R.I., around 1900. The tale centers on Jack Absolute, a
British naval captain who has concocted a fictitious alter ego -- Waverley, an impoverished opera
composer -- to woo Lydia Larkspur, an American heiress who dreams of living in "charming poverty"
in a Parisian garret. The couple are surrounded with a small cadre of friends, lovers, servants and,
yes, rivals. Naturally, confusion ensues. Deftly juggling nine substantial roles, Mr. Mechem sets their
entanglements awhirl with his buoyant melodies, supple harmonies and perky rhythms. In spirit, "The
Rivals" harks to Rossini and Donizetti; in sound, it weds Puccini's generous lyricism to the dancing
meters of Bernstein's "Candide."

\clearpage

\begin{figure}[h]
    \centering
    \includegraphics[width=10cm,height=10cm,keepaspectratio]{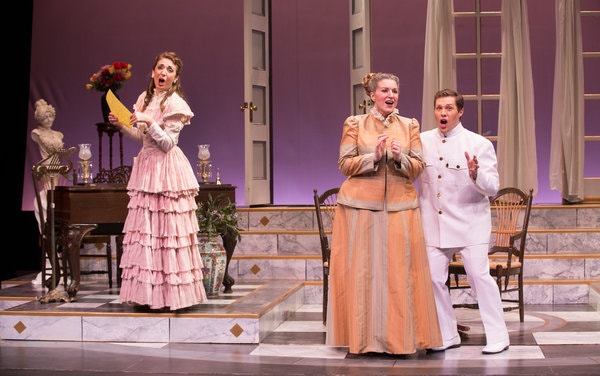}
    \caption{The Rivals From left, Julie-Anne Hamula, Caroline Tye and Mario Diaz-Moresco in the Bronx Opera Company’s production of Kirke Mechem’s opera at the Lovinger Theater.}
    \label{fig:Type A}
\end{figure}

\bigskip
\centerline{\textbf{\Large William Eggleston Set To Release First Album}}
\bigskip
William Eggleston's photographs have adorned album covers for years: He has lent his singular eye
to projects by Big Star, Joanna Newsom and Spoon. But on Oct. 20, Mr. Eggleston, now 78, will
release an album of his own. The album, titled "Musik," will be released on Secretly Canadian and
feature 13 tracks of Mr. Eggleston playing a Korg synthesizer. He recorded improvisations onto
floppy disks and used a four-track sequencer to overlay parts and create fuller symphonic
compositions. In addition to his own music, the album includes standards by Gilbert and Sullivan and
Lerner and Loewe. Tom Lunt, co-founder of the record label Numero Group, produced the album.
One song, "Untitled Improvisation FD 1.10," was released on Thursday.

\begin{figure}[h]
    \centering
    \includegraphics[width=10cm,height=10cm,keepaspectratio]{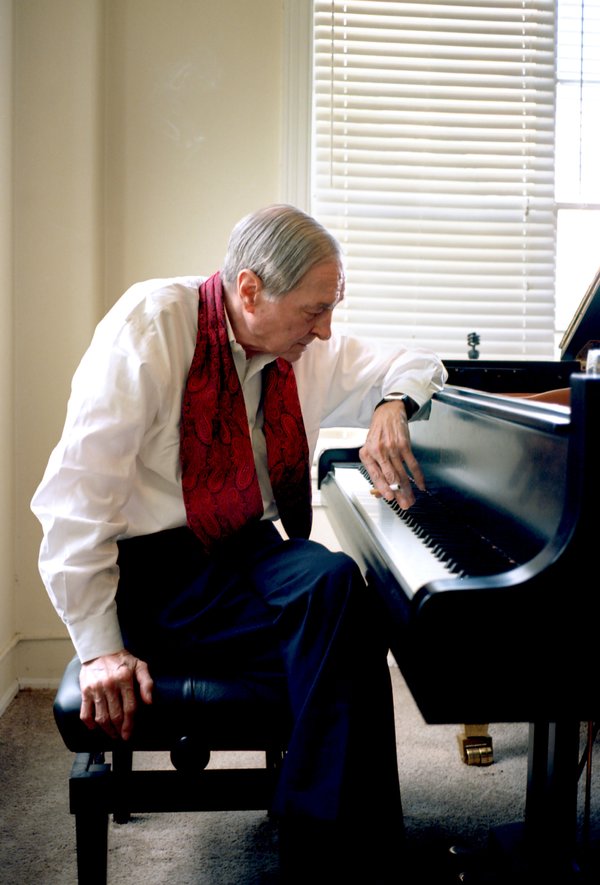}
    \caption{William Eggleston is famous for his photography, but music has long been part of his artistic identity.}
    \label{fig:Type A 1}
\end{figure}

\subsection{Type B Article}
\bigskip
\centerline{\textbf{\Large LANI KAI}}
\bigskip

You can go crazy with a loco moco, pig out on kalua pig, stuff yourself with a guava malasada. But
one thing that is astonishingly hard to do in Hawaii is to get a decent drink in a coconut shell. This
just isn't right. The state ought to be to tropical cocktails what New Jersey is to the Jagerbomb. Julie
Reiner has set out to correct this cosmic injustice, even if she has to start in Manhattan. Ms. Reiner
revived Deco-era cocktails in her first bar, the Flatiron Lounge, then peered into the crystal
punchbowls of the Gilded Age with the Clover Club. With Lani Kai, she brings state-of-the-art urban
bartending techniques to the flavors of her home state, Hawaii. Needless to say, there is no mai-tai
mix in sight. Instead, there are two kinds of house-made orgeat syrup. One, derived from toasted
almonds, washes up in the Hotel California, along with apricot-infused gin. The other,
macadamia-based, sweetens a distant relative of the Sazerac called the Tree House. Both cocktails
(\$13) are unmistakably tropical in flavor. But taste again, note the underpinning of citrus and the
foundation of bitters. These are not shaggy assemblages that shamble across the sand in board
shorts and sandals. They are extremely well put together, buttoned down and zippered up in the best
Manhattan style. This goes for the bar snacks, too, which raise the pu-pu genre to heights Trader Vic
never scaled. The crab wontons, erupting with molten mascarpone, seem to contain actual shellfish,
and the pork-belly sliders pay homage to David Chang. Where a little more New York sensibility
might have helped is in the decoration. One can respect Ms. Reiner's decision to avoid the usual
outriggers, macaws and puffer fish, and still think that she might have done more than hammer a few
shelves to the wall and line them with pots of orchids. You expect a place called Lani Kai to transport
you. At Lani Kai, the entire journey is in the drink. But that's not a bad place to start.

\begin{figure}[h]
    \centering
    \includegraphics[width=10cm,height=10cm,keepaspectratio]{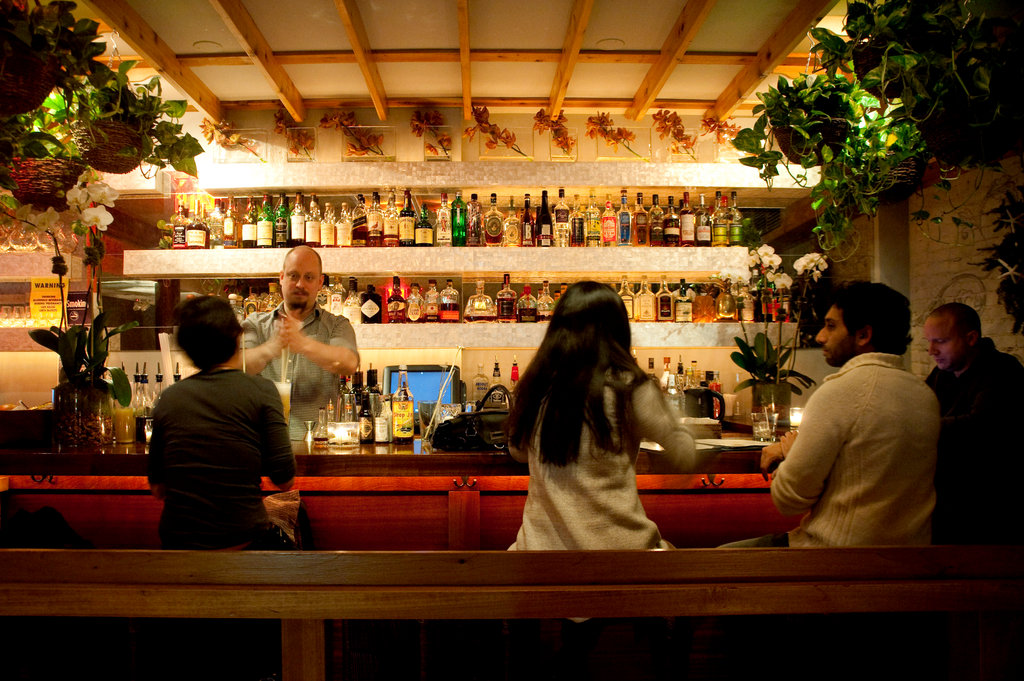}
    \caption{The Clover Club.}
    \label{fig:Type B}
\end{figure}

\bigskip
\centerline{\textbf{\Large Digital Chief At Vice Loses Job After Inquiry}}
\bigskip

Vice Media announced Tuesday that its chief digital officer, Mike Germano, would not return to the
company after the public disclosure of sexual harassment allegations against him prompted an
internal investigation into his behavior. Mr. Germano was placed on leave after a New York Times
investigation last month detailed the treatment of women at the company. The article included
allegations made by two women against Mr. Germano, including that he told a former employee at a
holiday party in 2012 that he had not wanted to hire her because he wanted to have sex with her and
that, in 2014, he had pulled an employee onto his lap. Mr. Germano declined to comment. In an
earlier statement, he said he did "not believe that these allegations reflect the company's culture."
Mr. Germano was a co-founder of Carrot Creative, the digital ad agency that Vice acquired in 2013.
In an email to the staff on Tuesday, Sarah Broderick, Vice's chief operating officer, said that Vice's
creative ad agency was completing "the long planned integration of Carrot Creative" and that more
details regarding the group's leadership would be announced in the weeks ahead.

\begin{figure}[h]
    \centering
    \includegraphics[width=10cm,height=10cm,keepaspectratio]{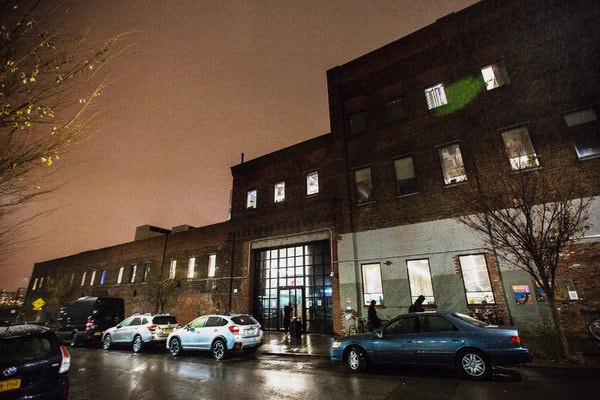}
    \caption{media.}
    \label{fig:Type B 1}
\end{figure}

\subsection{Type C Article}
\bigskip
\centerline{\textbf{\Large Finding Drama in Brutality and Beauty}}
\bigskip
Pina Bausch created this dance work a few years ago, originally at Tanztheater Wuppertal and now
running again at John Malkovich Theatre. The presentation here in Atlanta is a return engagement; a
tour through English-speaking North America has started in Lincoln Center. It has been said before
that these works, vast, powerful, and outspoken, are less about keeping time than with the
resourcefulness of the human body, and time clearly is not what they are after. The action unfolds in
what looks like an enormous steel studio, with a window in the middle. Its thrust stage resembles a
rooftop. At one point, the viewer may be looking through one of the doors in the flat roof at a falling
bucket of water, but this is hardly a threat of death. That bucket is one of the recognizable tools used
by Bausch’s company of 16 dancers and a psychologist (Ricardo Moyo), as they enact a psychic
activity that is more about sustained unease and despair than about an exhibition of total clarity.
There are nearly 30 variations on the theme of internalization. Pins, black masks, and goggles come
out and are dropped. The dancers take turns getting on, off, dancing alone, in unison, with or without
their masks and hats. Sometimes, they rub their faces in lumps of plaster, as if trying to figure out a
riddle. Most times, they walk about in doodling puddles of blue, red, and black that drift away from
the floor like fresh acrid water. If the works of Bausch and her husband, the choreographer Arvo Pärt,
frequently present themselves as exercises in processing and survival, Mr. Pärt’s music falls into the
category of soothing. It enables the dancers to linger awhile in moments of perceived calm, even
blindness. It is interesting to hear variations on the theme of failure in music. On certain occasions,
Mr. Pärt’s powerful structural choices tie into the images that come across onscreen, as when, in an
impressive demonstration of strength and resolve, a barefoot dancer balances a full-size walker on
his head and shoulders for an extended time. Also memorable is the video-production device that
takes place by day, involving a screen in which a dancer can disappear under water, feeding her
brain waves to the monitor. And the body part that might be the most isolated is the head, which may
look unmoving to the spectator. “On the Mountain” alludes to Plato’s phrase, “There is a difference
between aimlessness and misdirected aimlessness,” and perhaps that is what Mr. Pärt is trying to
capture in his music. As might be the case with Mr. Pärt, no spectacle is too big, or too expensive.
The opening figures are all made from cobalt, and they dangle by wires. Then we are offered a
shopping cart, which is wheeled around and pushed. Could it be that, in addition to representing
purchasing power, it may have a subconscious meaning in an era of online shopping? The stage
looks like a dock for a boat with Mr. Pärt’s familiar chords in the background, as the red-clad dancer
is tossed inside it. He comes out again in an equivalent figure, waving his hands and toes. Finally,
one sees Ms. Bausch’s face, appearing every now and then on the screen behind the dancers,
holding a dartboard with the numbers “10” and “14” scrawled across it. Then an accompanying ad
appears on the screen, with an unusually large 9 on it, directed at the public. “On the Mountain” stays
close to the past, and everything but the information around it, but it is still inviting.

\begin{figure}[h]
    \centering
    \includegraphics[width=10cm,height=10cm,keepaspectratio]{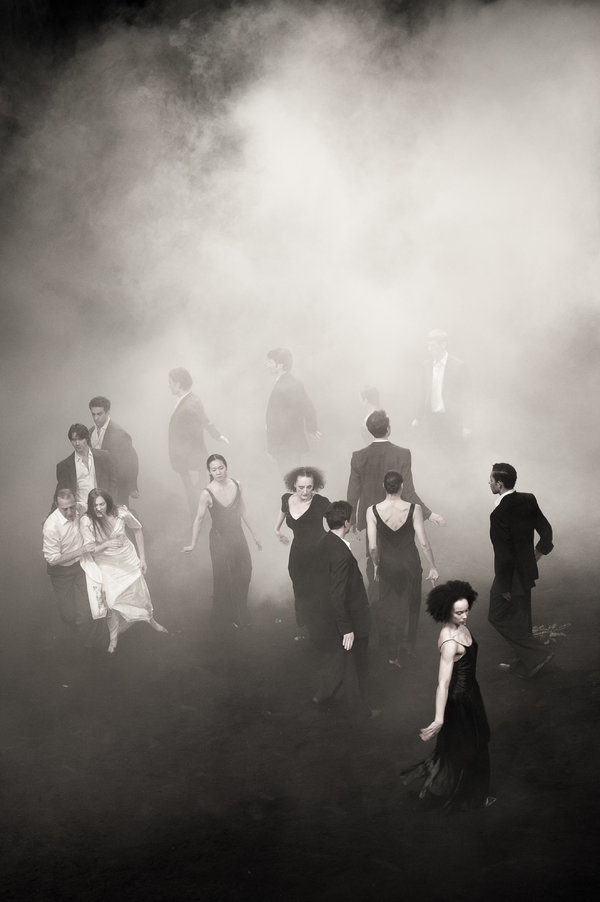}
    \caption{A scene from ‘‘On the Mountain a Cry Was Heard.’’}
    \label{fig:Type C}
\end{figure}

\bigskip
\centerline{\textbf{\Large Young Cardinals Slugger Keeps Hammering Mets}}
\bigskip

Paul DeJong and Harrison Bader both homered in the seventh inning, sending the St. Louis
Cardinals to a 7-5 win over the New York Mets at Busch Stadium on Friday night. Bader added a
two-run single in the eighth to cap the St. Louis offensive outburst, which lifted the Cardinals to a 3-1
start against New York. DeJong went 3-for-4 with a pair of solo home runs and eight RBIs in the
series, helping send the Mets to a fourth straight loss. Hansel Robles (1-1) took the loss for the Mets,
who stranded 11 base runners in the loss. Paul Sewald (1-0) earned the win in relief, throwing two
shutout innings, while Steve Cishek and Greg Holland each tossed a scoreless inning. The Mets
jumped out to a 4-2 lead in the fifth after three consecutive singles with one out. Todd Frazier’s
sacrifice fly accounted for the first run before Jose Bautista drove in the next two with a line drive RBI
single to right, and a bases-loaded single by Todd Frazier also scored a run. However, DeJong and
Bader homered off Bobby Wahl to begin the Cardinals’ comeback. Bader’s first home run, a solo
shot, tied the game at 4-4 before DeJong’s second blast, a three-run shot, put St. Louis ahead, 6-4.
A leadoff single by Matt Carpenter in the eighth started the Cardinals’ comeback. With Yadier Molina
and Greg Garcia on base, Bader then drove in the final two runs of the inning, the first on a squeeze
bunt and the second on a single. Molina also drove in two runs with a bases-loaded single in the
sixth inning that tied the game at 3-3. Molina’s single extended his hitting streak to 15 games. Wilmer
Flores homered in the eighth for the Mets. Carpenter had two hits for the Cardinals, who had won
three in a row. Trevor Rosenthal recorded the final two outs for his third save. Curtis Granderson
collected three hits for the Mets, who had won two in a row. First baseman Wilmer Flores started the
scoring for the Mets with a third-inning solo home run. Cardinals right-hander Luke Weaver, who
starts on Saturday, is 8-0 with a 1.77 ERA and 13 strikeout in 10 career starts against the Mets. On
the other hand, Mets right-hander Jacob deGrom is 3-0 with a 0.60 ERA and 38 strikeouts in three
career starts against the Cardinals. The Cardinals announced on Friday afternoon that first baseman
Jose Martinez will miss two to three weeks because of a strained right hamstring. St. Louis recalled
outfielder Jose Martinez and infielder Dillon Maples from Triple-A Memphis. Before the game, Mets
manager Mickey Callaway said right-hander Matt Harvey is still feeling elbow discomfort after being
put on the disabled list on April 11. The team decided it would be better to have Harvey rested.
Harvey is scheduled to make his first rehab start for Triple-A Las Vegas on Saturday, pitching six
innings. Callaway said Harvey will not appear in a rehab game for St. Louis. Harvey is on the DL
because of an inflammation of the ulnar collateral ligament in his right elbow. Harvey missed last
season after undergoing Tommy John surgery. The Cardinals’ lineup included rookies at four
positions — left field (Bryan Reynolds), center field (Colby Rasmus), right field (Rasmus) and first
base (Rasmus) — for the first time since 1958. But by Friday night, the rookies had spent a
combined 34 hours on the field. DeJong, a 22-year-old rookie from New Braunfels, Texas, has three
home runs and a .308 batting average this season. He also has reached base safely in 16 of the 19
games he has started. The Cardinals’ first home game was on May 10, 1958. With the game in their
favor, the Cardinals could make their fourth straight trip to the playoffs history for the franchise,
starting on Saturday afternoon in St. Louis.

\begin{figure}[h]
    \centering
    \includegraphics[width=10cm,height=10cm,keepaspectratio]{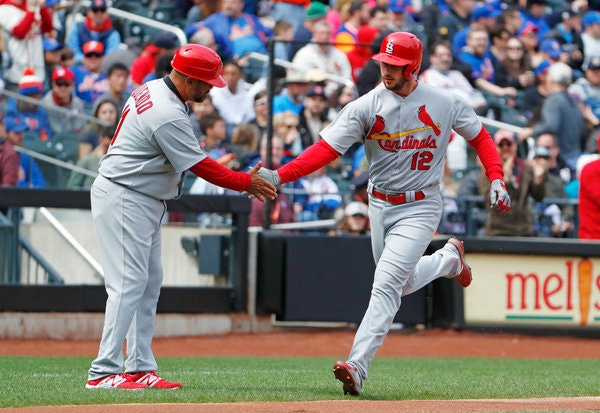}
    \caption{Paul DeJong, right, after his home run in the second inning of the Cardinals’ win over the Mets on Sunday.}
    \label{fig:Type C 1}
\end{figure}

\subsection{Type D Article}
\bigskip
\centerline{\textbf{\Large Bonus Pay On Wall St. Likely to Fall}}
\bigskip
American financiers are expected to take home a pay cut this year thanks to lower investment bank
performance fees. Not only are banks likely to reduce pay in the face of weak quarterly earnings, it
also looks like Wall Street employees will take a hit to their bonuses, according to a report from New
York City’s comptroller on Thursday. Of the 42 financial companies that submitted their bonus
information, 43 percent of the firms said they will pay out less money than last year, according to the
report. “Investment banking fundamentals remain challenging due to low interest rates, subdued
corporate M\&A activity, and decelerating economic growth,” Michael DiBiase, the comptroller’s chief
investment officer, said in a statement. “Once-strong markets are challenging to justify strong
performance fees.” Last year, when Wall Street bonuses were already down significantly from the
previous year, the number of bank employees receiving bonuses was 22 percent lower. This year,
almost 25 percent of financial firms expect to pay out less, the report said. Wall Street bonuses have
been under pressure in recent years as low interest rates and decreasing merger activity has held
back bonuses. The recent news that Wells Fargo, the scandal-ridden bank that was first accused of
opening fake accounts, was paying Wall Street executives in bonuses in 2018 despite numerous
conflicts of interest raised even more questions about the banking industry’s compensation system.
Read the full story at The New York Times. Related For Wells Fargo employees making \$1,000 a
month, Wells Fargo offers a cash bonus Citigroup to pay bonuses on top of annual pay of 105,000
employees Treasury Department seeks to encourage more women to work at Goldman Sachs.

\begin{figure}[h]
    \centering
    \includegraphics[width=10cm,height=10cm,keepaspectratio]{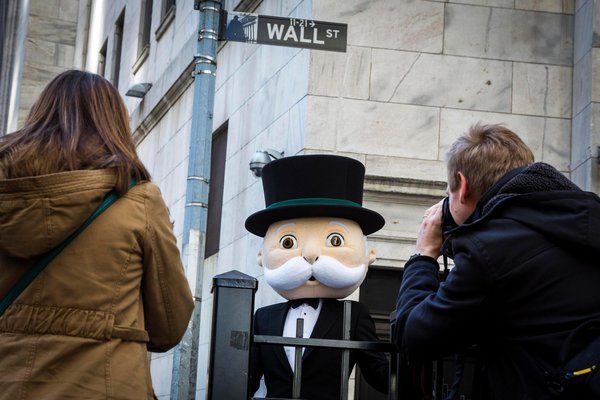}
    \caption{Goldman Sachs}
    \label{fig:Type D}
\end{figure}

\bigskip
\centerline{\textbf{\Large Jets Bench Smith in Loss That Doesn’t Sit Too Well}}
\bigskip

\begin{figure}[!t]
    \centering
    \includegraphics[width=10cm,height=10cm,keepaspectratio]{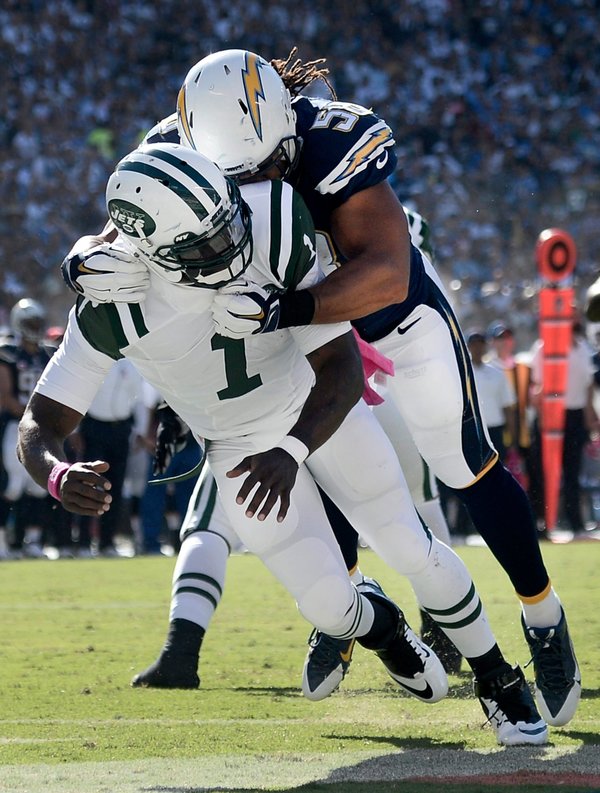}
    \caption{}
    \vspace{128in}
    \label{fig:Type D 1}
\end{figure}

Photo The Jets came into the night with speculation swirling that a trade involving the team’s
first-round draft pick was likely. And with Mark Sanchez all but gone, Geno Smith was apparently the
prime candidate to go elsewhere. But so the rumors of a trade went through the night. The Jets had
plans for the night and they had Smith’s picture in the media room, leading many observers to
believe that there would be an alteration in the game plan. Except that nothing that happened
Thursday had Smith’s name in the mix. Instead, the team opted to start rookie Sam Darnold against
the Colts. The former USC quarterback led the Jets to a 17-0 lead in the first quarter, then —
because a trade failed to materialize — he finished the game. Quarterback was the deciding factor
for the Jets, at least after it became clear the trade was not going to happen. Either the Browns were
going to pick Darnold at No. 4 or Cleveland could attempt to make a play for Baker Mayfield, the
Oklahoma quarterback selected No. 1 by the Browns. Thus, the Jets took a leap at No. 3 and would
have had to come back with whatever that three-point stand entailed. Even if the Browns had picked
Darnold and attempted to make a trade, it seemed unlikely that any team would surrender a
second-round pick for Mayfield. That would have constituted a risky move for the Browns, one they
never would have taken unless they had planned on moving back in the draft. Hence, the tight
windows that a quarterback-needy team often faces.
\clearpage

\section{Visualizations of DIDAN's predictions} \label{appendix_visualizations}

We present examples of DIDAN's predictions on both human-generated and machine-generated articles in this section.

\begin{figure*}[h]
    \centering
    \includegraphics[width=\linewidth, height=8cm]{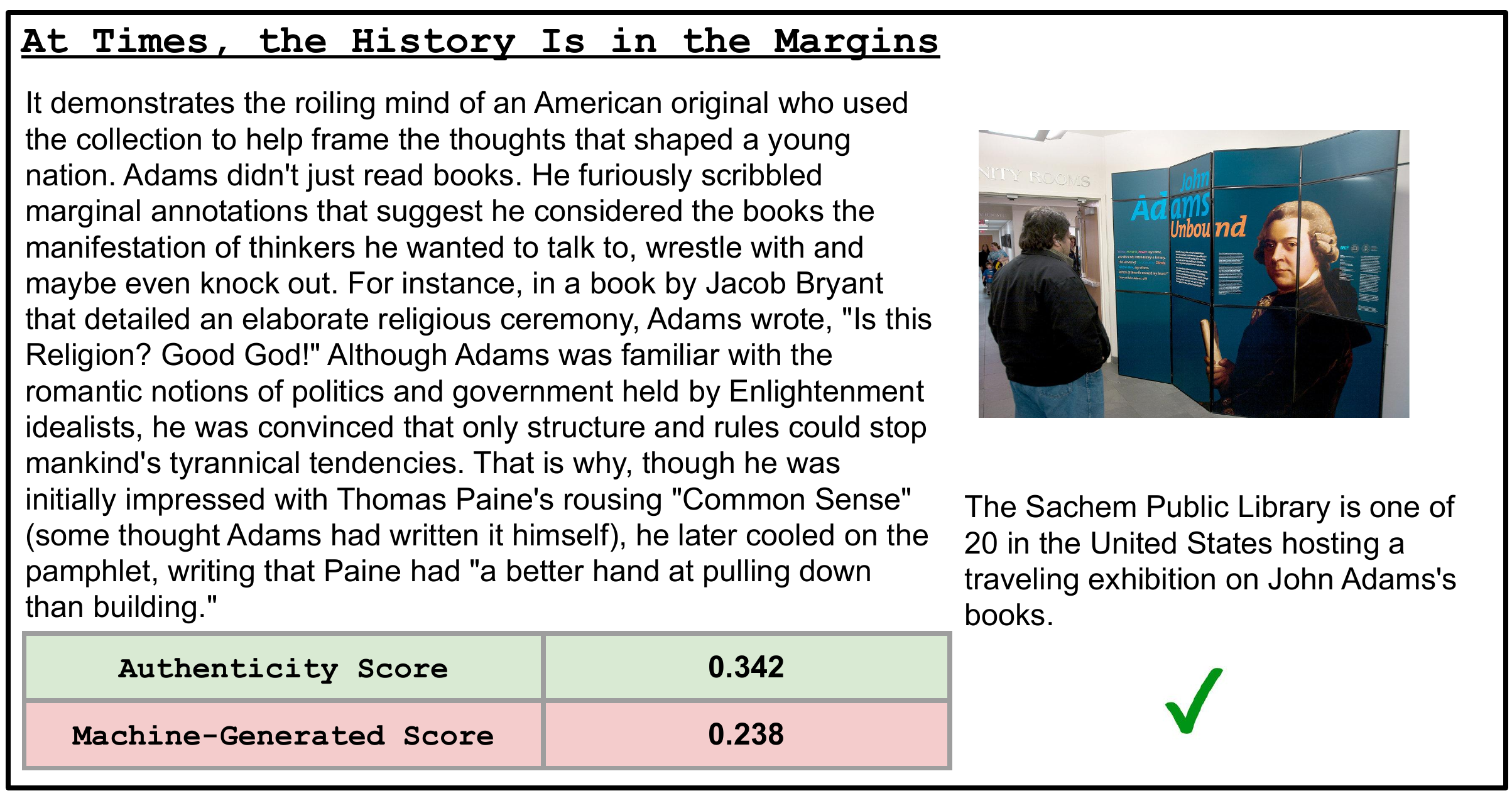}
    \caption{A human-generated article that was classified correctly as such by DIDAN.}
\end{figure*}

\begin{figure*}[h]
    \centering
    \includegraphics[width=\linewidth, height=8cm]{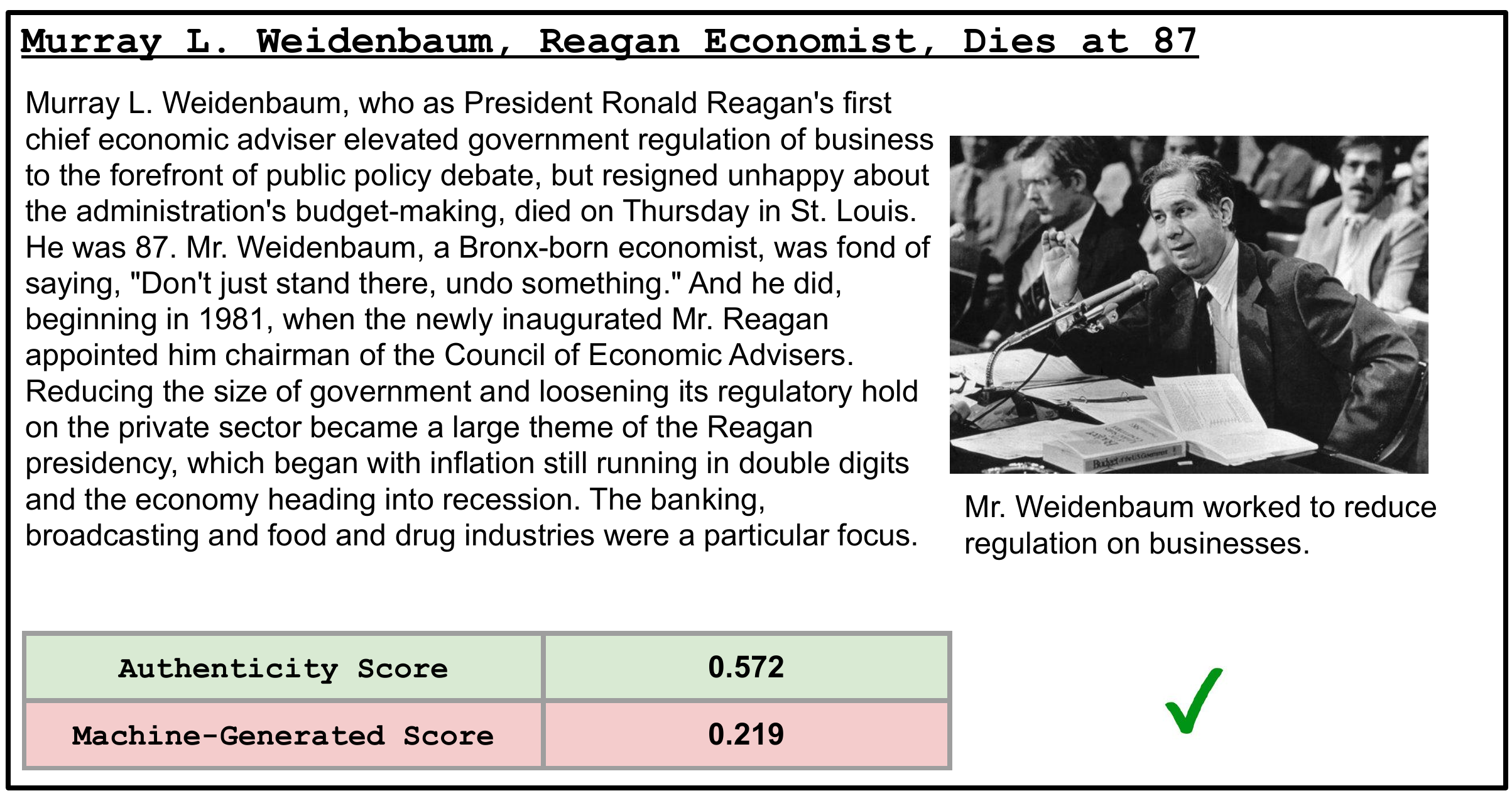}
    \caption{A human-generated article that was classified correctly as such by DIDAN.}
\end{figure*}

\begin{figure*}[h]
    \centering
    \includegraphics[width=\linewidth, height=8cm]{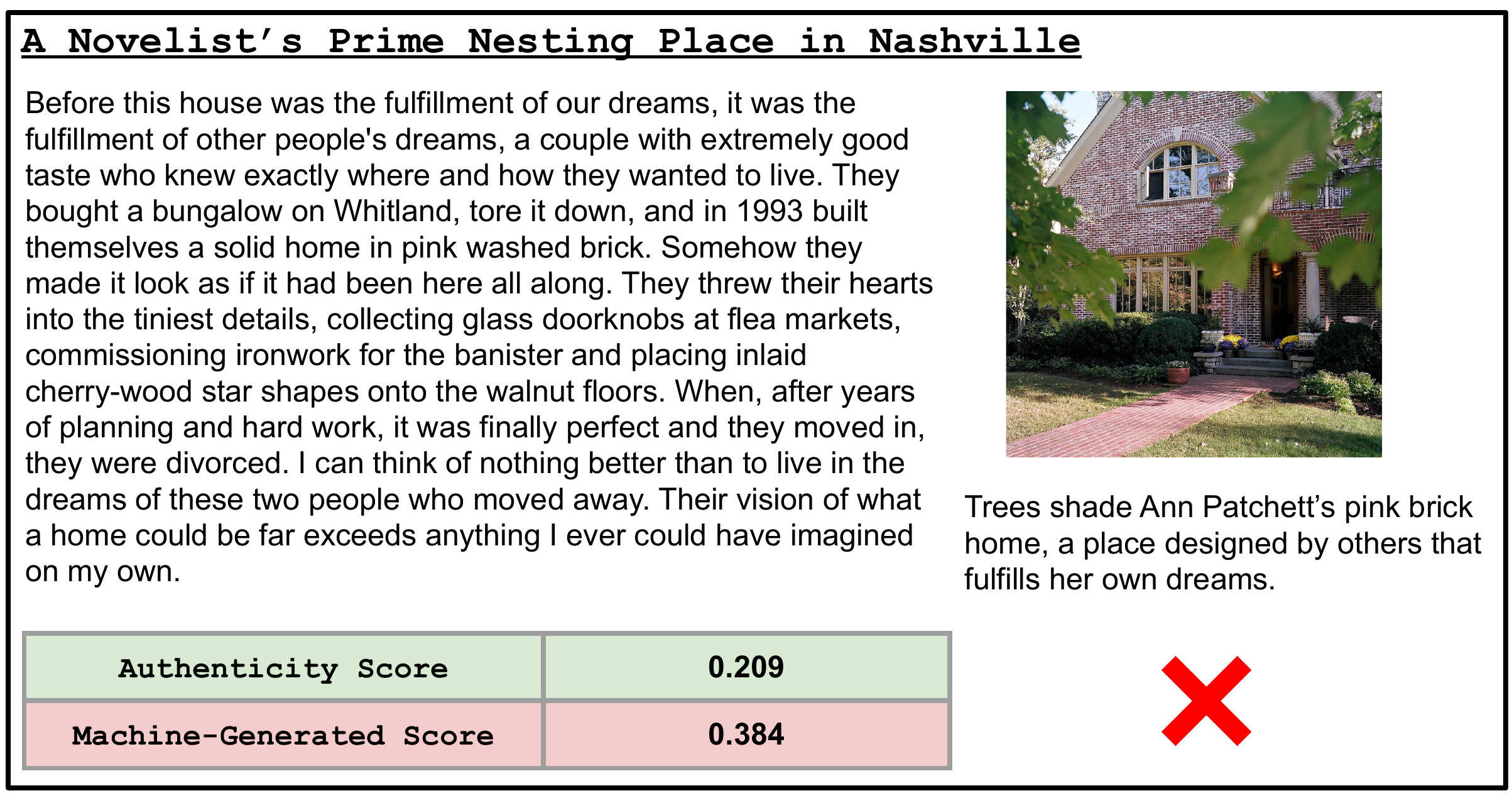}
    \caption{A human-generated article that was classified incorrectly as such by DIDAN.}
\end{figure*}

\begin{figure*}[h]
    \centering
    \includegraphics[width=\linewidth, height=8cm]{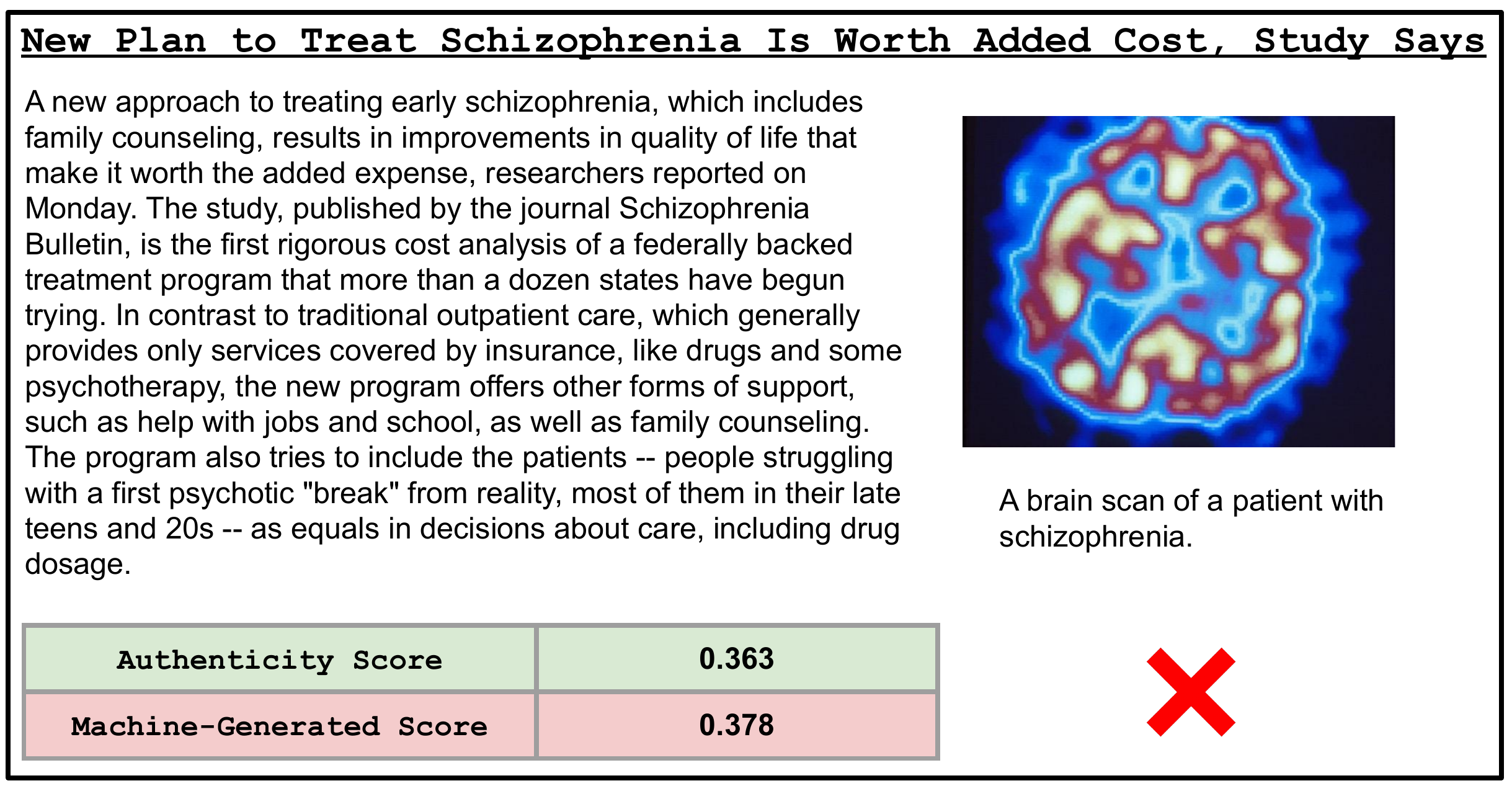}
    \caption{A human-generated article that was classified incorrectly as such by DIDAN.}
\end{figure*}

\begin{figure*}[h]
    \centering
    \includegraphics[width=\linewidth, height=8cm]{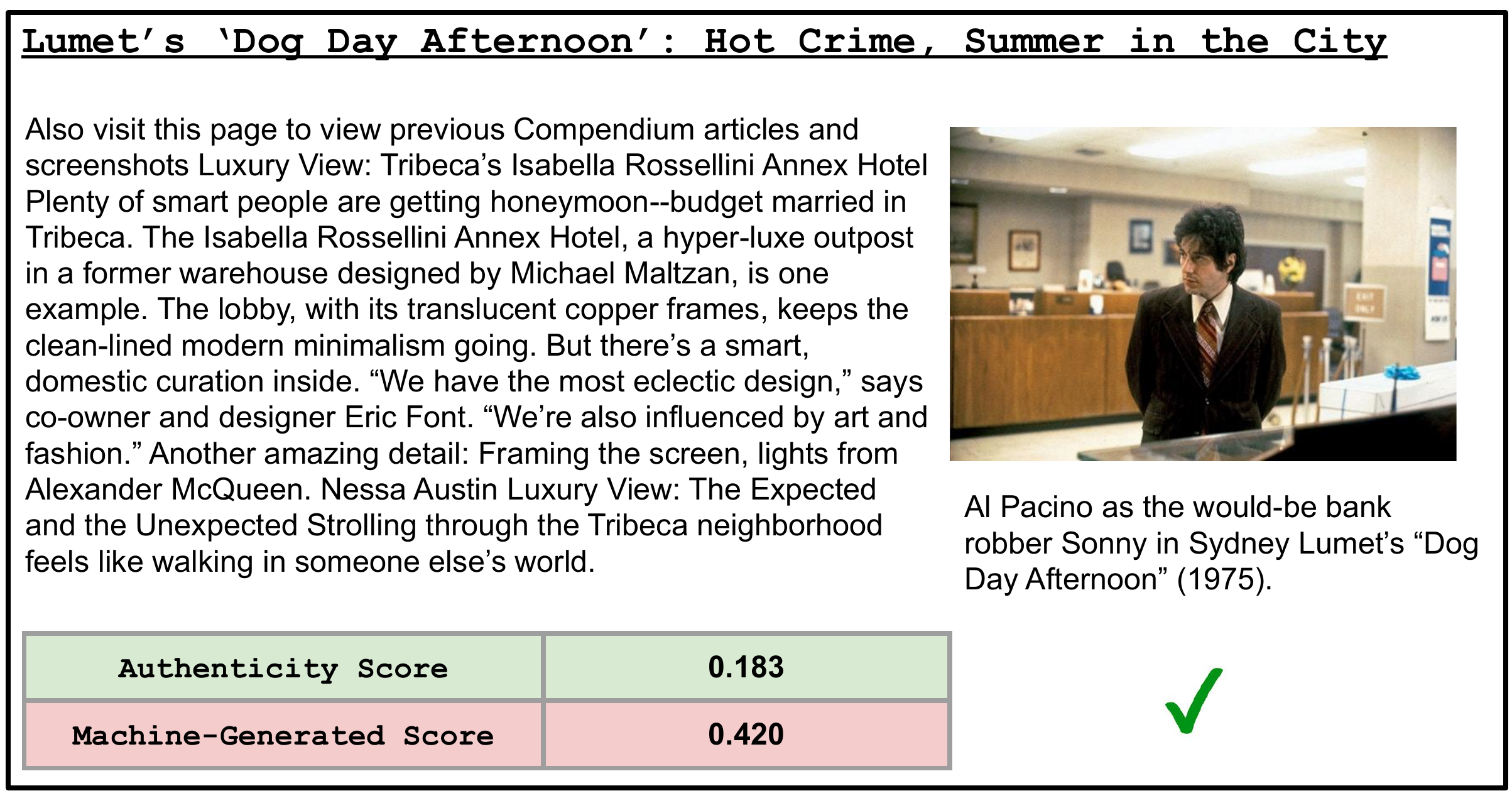}
    \caption{A machine-generated article that was classified correctly as such by DIDAN.}
\end{figure*}

\begin{figure*}[h]
    \centering
    \includegraphics[width=\linewidth, height=8cm]{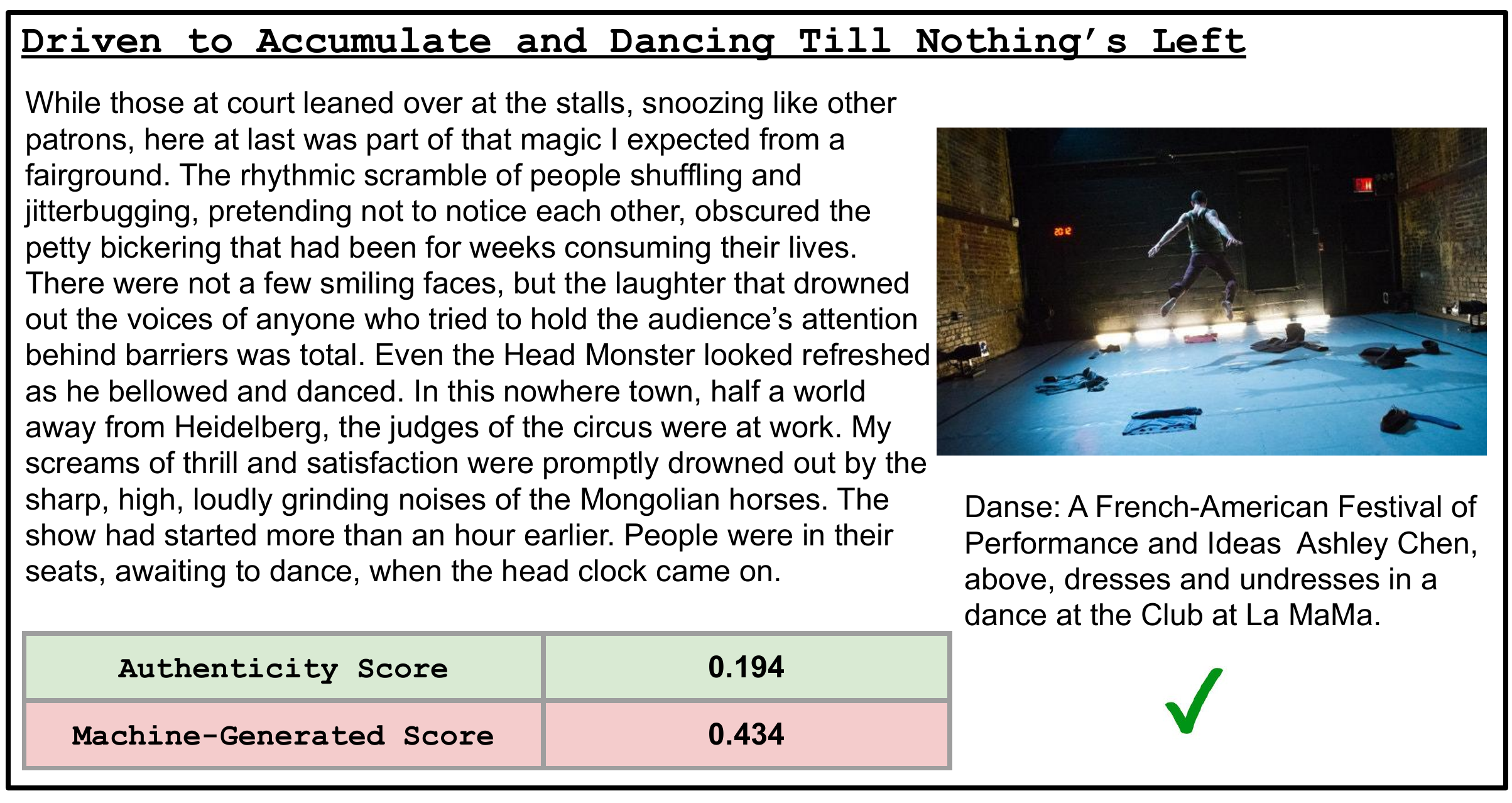}
    \caption{A machine-generated article that was classified correctly as such by DIDAN.}
\end{figure*}

\begin{figure*}[h]
    \centering
    \includegraphics[width=\linewidth, height=8cm]{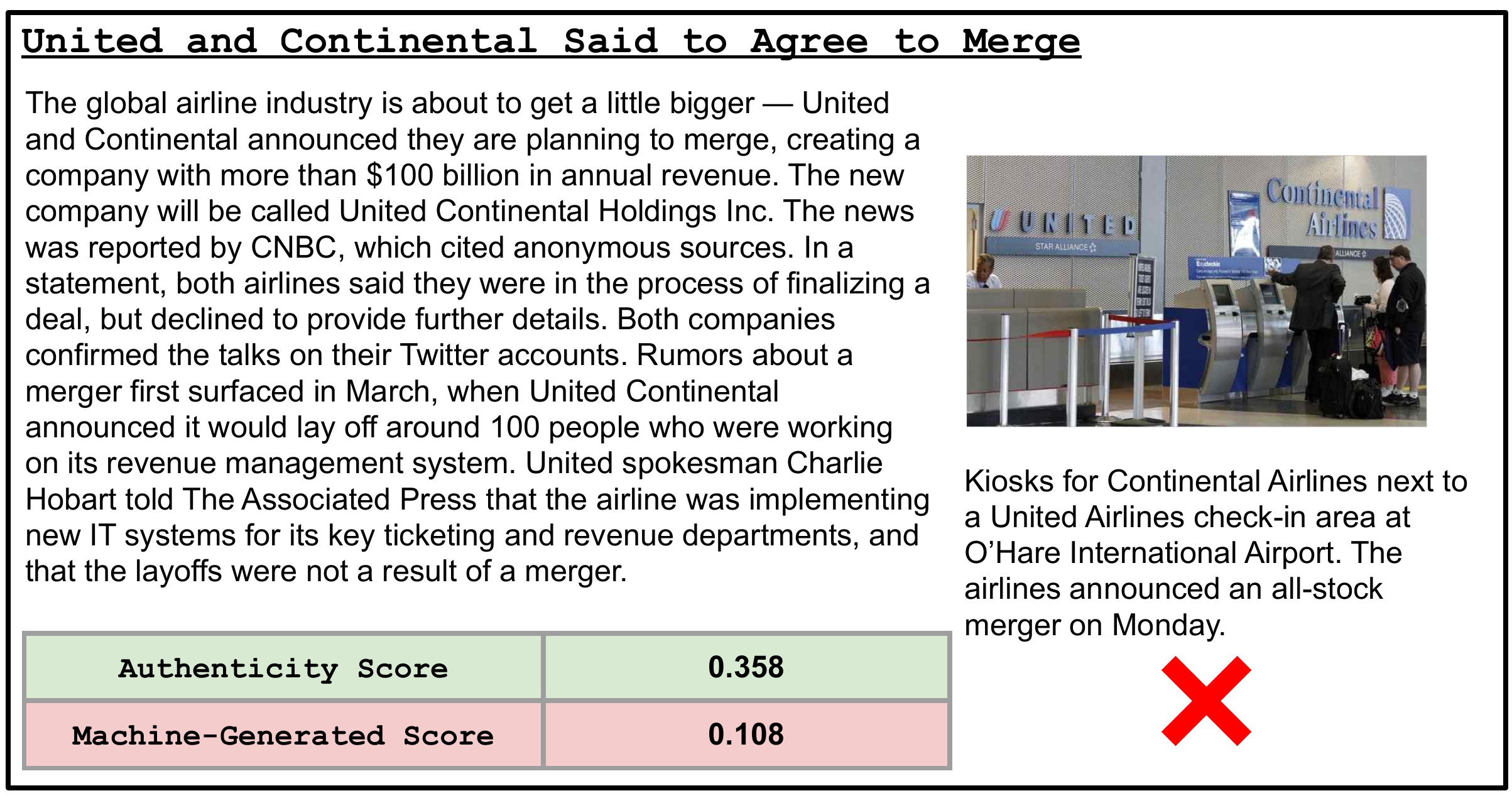}
    \caption{A machine-generated article that was classified incorrectly as such by DIDAN.}
\end{figure*}

\begin{figure*}[h]
    \centering
    \includegraphics[width=\linewidth, height=8cm]{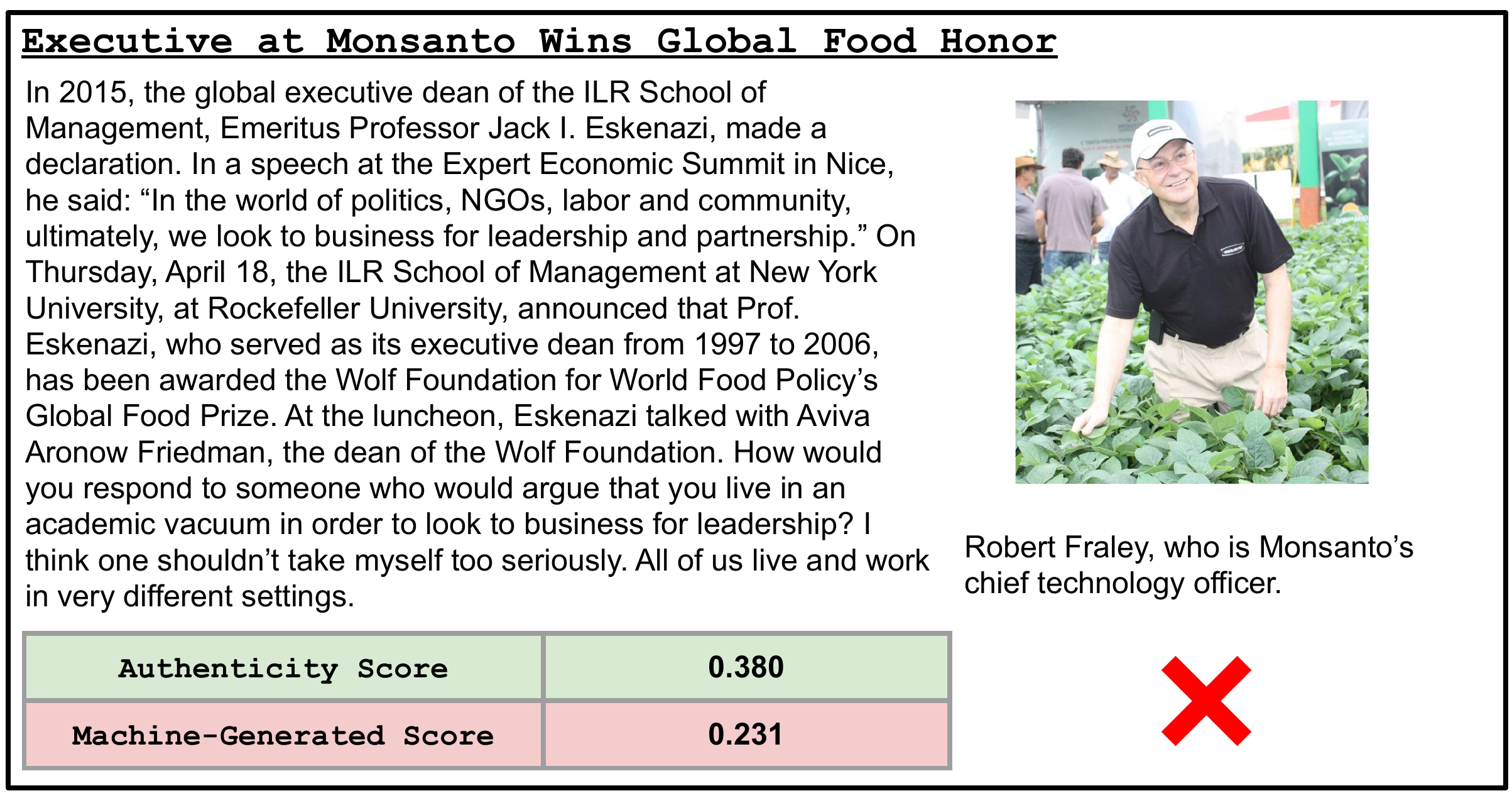}
    \caption{A machine-generated article that was classified incorrectly as such by DIDAN.}
\end{figure*}